\RequirePackage{fix-cm}
\documentclass[twocolumn]{svjour3}

\smartqed
\usepackage{titlesec} 
\titleformat{\paragraph}[runin]
{\normalfont\bfseries}{\theparagraph}{1em}{}
\titlespacing{\paragraph}{0em}{1ex}{.5em}
\usepackage{todonotes}
\usepackage{bm}
\usepackage[table]{xcolor}
\usepackage{graphicx}
\usepackage{newtxtext, newtxmath}
\usepackage{balance}
\usepackage{amsmath}
\usepackage[colorlinks=true, linkcolor=blue, citecolor=blue, urlcolor=blue]{hyperref}
\usepackage{xcolor}
\usepackage{latexsym}
\usepackage{makecell}
\usepackage{diagbox}
\usepackage{adjustbox}
\usepackage{caption}
\usepackage{soul}

\usepackage{booktabs}
\usepackage{multirow}
\usepackage{stfloats}
\usepackage{threeparttable}
\usepackage{url}
\usepackage{array}
\usepackage{ragged2e}
\usepackage{enumitem}
\usepackage{microtype}
\usepackage{pifont}
\usepackage{algorithm}
\usepackage{algorithmic}
\usepackage{subfig}

\usepackage{enumerate}

\usepackage[round, sort]{natbib}
\usepackage{xcolor}
\newcommand{\todooo}[1]{\textcolor{red}{#1}}  
\newcommand{\todoo}[1]{#1}  
\renewcommand{\todooo}[1]{#1}

\begin{document}
\sloppy

\title{Accurate and Efficient Object Pose Estimation via the Aggregation of Diffusion Features\thanks{\textit{Corresponding author: Hongguang Wang.}}}

\author{Tianfu Wang \and  Guosheng Hu \and Hongguang Wang
}

\institute{
    Tianfu Wang$^{1, 2}$ \at
    \email{wangtianfu100@gmail.com}         		
    \and
    Guosheng Hu$^{3}$ \at
    \email{huguosheng100@gmail.com}
    \and
    Hongguang Wang$^{1}$ \at
    \email{hgwang@sia.cn}
    \\	
    \and
    $^{1}$ State Key Laboratory of Robotics, Shenyang Institute of Automation, Chinese Academy of Sciences\\
    $^{2}$ University of Chinese Academy of Sciences, Beijing, 100049, China\\
    $^{3}$ University of Bristol, Bristol, UK
}	

\date{Received: / Accepted:}

\maketitle

\begin{abstract}

Estimating the pose of objects from images is a crucial task of 3D scene understanding, and  recent approaches have shown promising results on very large benchmarks. 
However, these methods experience a significant performance drop when dealing with unseen objects. 
We believe that it results from the limited generalizability of image features.
To address this problem, we have an in-depth analysis on the  features of diffusion models, e.g. Stable Diffusion, which hold substantial potential for modeling unseen objects.
Based on this analysis, we then innovatively introduce these diffusion features for object pose estimation. 
To verify the efficacy of diffusion features for object pose estimation, we propose three distinct architectures (vanilla, nonlinear, and context-aware weight aggregations) that capture and aggregate diffusion features for comparative analysis. The results demonstrate strong generalization performance in object pose estimation. In real applications, generalizability to unseen data is important, on the other hand, the efficiency is also crucial. To achieve an efficient feature aggregation, we propose a confidence adaptive aggregation network that 
automatically selects the discriminative features rather than uses all the features, achieving a better speed-and-accuracy trade-off.
\todoo{Our method achieves substantial improvements on LM, O-LM, and T-LESS, and also demonstrates measurable gains on the seven core datasets of the BOP Challenge.}
In particular, our confidence adaptive aggregation network  achieves higher accuracy than the previous best arts on \emph{unseen} objects: 97.7\% vs. 93.5\% on Unseen LM, 85.5\% vs. 76.3\% on Unseen O-LM, showing the strong generalizability of our method. 
\todoo{On the large-scale BOP benchmark, our method also provides measurable gains, with an average recall of 58.3 compared to 57.9 previously.
In addition, CAA reduces computational cost by 1.3–1.5× compared to the CWA variant while maintaining comparable accuracy. Furthermore, CAA reaches real-time performance, achieving over 68 FPS and offering a substantially improved accuracy-efficiency trade-off.}

\end{abstract}

\section{Introduction} 
\label{sec:intro}
\begin{figure*}[!t]
\centering
\footnotesize
\includegraphics[width=0.99\textwidth]{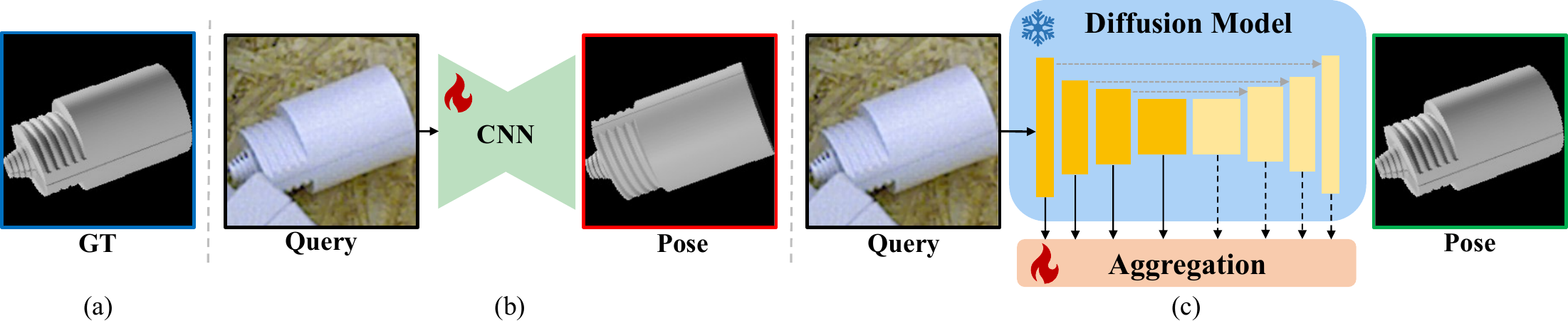}
\caption{
Unseen object  pose estimation  of  one state-of-the-art method~\citep{nguyen2022templates} and our method.
(a) We render an image using the ground truth pose of an unseen object.
(b) Template-pose~\citep{nguyen2022templates} learns image features by fine-tuning a self-supervised learning~\citep{he2020momentum} pre-trained model.
(c) Our method aggregrates features of different granularity from a diffusion model to achieve better pose estimation than template-pose~\citep{nguyen2022templates} on the unseen object. {The snowflake and flame symbols represent `parameters frozen’ and `fine-tune’, respectively.}}
\label{fig:intro}
\end{figure*}

Estimating the pose of objects from images is crucial for various real-world applications, including robot manipulation~\citep{collet2011moped,tremblay2018deep,zhu2014single}, augmented reality~\citep{marchand2015pose}, and many others. In these contexts, achieving accurate and reliable pose estimation is important. 
However, real-time performance is equally vital, as delays can adversely affect system effectiveness and user experience. 
The ability to quickly respond while maintaining accurate estimations is essential for enabling intelligent interactions. 
Fortunately, advancements in deep learning have significantly enhanced the performance of object pose estimation.

The most popular methods for object pose estimation can be  categorized into three approaches: indirect methods~\citep{rad2017bb8, park2019pix2pose, li2019cdpn, zakharov2019dpod, hodan2020epos, haugaard2022surfemb}, direct methods~\citep{xiang2017posecnn, manhardt2019explaining, do2019real, kehl2017ssd, wang2021gdr, chen2022epro}, and template-based methods~\citep{wohlhart2015learning, balntas2017pose, sundermeyer2018implicit, li2018deepim, labbe2020cosypose}. Template-based methods involve matching the input image of an object to a set of template images with various poses rendered by 3D models to achieve the object's pose. In this work, we focus on template-based methods due to their simplicity, increasing popularity, and promising accuracy.

Despite these efforts, existing approaches still face notable limitations when generalizing to unseen objects. 
Recent template-based methods such as~\citep{sundermeyer2020multi,nguyen2022templates,zhao2022fusing} show promising results, 
but their performance drops substantially on unseen objects. For example, Template-Pose~\citep{nguyen2022templates} achieves 99.1\% accuracy on the seen LM dataset 
but only 94.4\% on the \emph{unseen} LM dataset. 
\todoo{More recent approaches such as MegaPose~\citep{labbe2022megapose} demonstrate that large-scale synthetic training can improve generalization to unseen objects. GigaPose~\citep{nguyen2024gigapose} extends this idea with a very large DINO-based encoder and hundreds of millions of trainable parameters, achieving strong performance at substantially higher computational cost. GenFlow~\citep{moon2024genflow} refines poses by predicting optical flow between rendered and observed images. Training-free methods such as FoundPose~\citep{ornek2024foundpose} rely entirely on frozen pretrained features for pose estimation. However, its performance remains limited, attaining 56.8 average recall with five hypotheses, which is below the 57.9 achieved by GigaPose. These observations highlight the need for representations that retain the generalizability of pretrained features while still supporting lightweight task-specific adaptation.}

The performance limitations of existing template-based methods~\citep{sundermeyer2020multi,nguyen2022templates,zhao2022fusing} largely stem from the restricted discriminative strength of their learned feature representations, as further discussed in Sect.~\ref{subsec3-2}.
\todoo{Large pretrained models~\citep{labbe2022megapose,moon2024genflow,nguyen2024gigapose} generalize well but are costly, while training-free methods~\citep{ornek2024foundpose} are efficient but not adaptable. This motivates representations that keep pretrained generalization yet allow minimal adaptation for pose estimation.}
Recent studies show that text-to-image diffusion models produce robust intermediate features with strong generalizability~\citep{zhan2023does, xu2023open, tang2023emergent, luo2023diffusion, li2023guiding}. 
This promising generalizability can be attributed to four main factors: 
1) the rich semantics introduced by text supervision, 
2) the multi-timestep encoding that captures diverse levels of granularity, 
3) the capability to encode 3D-related cues such as geometry and depth, 
and 4) the use of large-scale, diverse training data exemplified by Stable Diffusion~\citep{rombach2022high}. 
Since these properties collectively indicate that diffusion features already possess strong discriminative and generalizable characteristics, they form a compelling foundation for unseen-object pose estimation. 
This motivates us to investigate how such features can be effectively adapted and aggregated to form representations suitable for object pose estimation, while maintaining practical training efficiency, as illustrated in Fig.~\ref{fig:intro}.


To verify the efficacy of diffusion features for object pose estimation, we conduct an in-depth analysis of the intermediate features of diffusion models. We observe that features from different layers of the diffusion model capture various dynamics. Based on this, we propose three architectures to aggregate all diffusion features into an optimal feature for object pose estimation. First, we design a Vanilla Aggregation (VA) network which aligns the features to the same dimension through linear mapping and then aggregates them via element-wise addition. However, the limited capability of this  linear network constrains its performance. To enhance the network's fitting capability, we introduce Nonlinear Aggregation (NA) network, which replaces the linear mapping with a bottleneck module consisting of three convolution layers and ReLU functions. Finally, to obtain the optimal weights for different features, we design a Context-aware Weight Aggregation (CWA) network, which learns the weights based on the context.

In real applications, both generalizability to unseen objects and computational efficiency are critical. We observe that the intermediate diffusion model features capture various dynamics, as illustrated in Fig. \ref{fig:feat}.
Motivated by this, we propose a network called Confidence Adaptive Aggregation (CAA), which 
adaptively selects a subset of all the features 
based on the input, greatly reducing the computations. At the core of this approach is a dynamic aggregation module that flexibly aggregates a variable number of features, overcoming the limitations of a fixed feature count. We also introduce a confidence prediction module that assigns a confidence score to the aggregated features, allowing the model to dynamically decide when to stop the aggregation process. This early exit mechanism ensures an optimal trade-off between computational efficiency and accuracy. By halting feature aggregation once an optimal confidence threshold is reached, the network achieves efficient feature selection while maintaining strong discriminative power.


To verify the efficacy of our aggregation methods, we first conduct experiments on three popular benchmarks: LINEMOD (LM)~\citep{hinterstoisser2013model}, Occlusion-LINEMOD (O-LM)~\citep{brachmann2014learning}, and T-LESS~\citep{hodan2017t}. On these datasets, our network achieves superior accuracy compared with state-of-the-art methods and significantly reduces the performance gap between seen and \emph{unseen} objects. Second, we examine the efficiency of our design during inference. The proposed CAA network adaptively determines the number of diffusion features to aggregate, which substantially reduces computation while maintaining high accuracy. \todoo{Finally, we evaluate our method on the BOP Challenge~\citep{sundermeyer2023bop} to assess its training efficiency. Even with substantially fewer trainable parameters, our method attains competitive performance, demonstrating that strong generalization can be achieved without heavy training cost.}



In summary, this work has the following contributions:


\begin{itemize}[leftmargin=20pt]

\item We have an in-depth analysis of the diffusion features, which exhibit great potential for modeling \emph{unseen} objects, and creatively incorporate diffusion features into object pose estimation. To our knowledge, we are the first to innovatively and systematically investigate the aggregation of diffusion features for object pose estimation.
\item We  propose three aggregation networks which can effectively capture different dynamics of the diffusion features, leading to the promising generalizability of object pose estimation.
\item We propose a Confidence Adaptive Aggregation (CAA) network that dynamically adjusts feature aggregation based on a confidence score, integrating an early exit mechanism to halt aggregation once sufficient confidence is achieved. To the best of our knowledge, this is the first application of an early exit mechanism in object pose estimation.
\item Our approach greatly outperforms the state-of-the-art methods on three benchmark datasets, LINEMOD (LM)~\citep{hinterstoisser2013model}, Occlusion-LINEMOD (O-LM)~\citep{brachmann2014learning}, and T-LESS~\citep{hodan2017t}. 
In particular, our method performs significantly better than other methods on \emph{unseen} datasets, 97.7\% vs. 93.5\%~\citep{nguyen2022templates} on Unseen LM, 85.5\% vs. 76.3\%~\citep{nguyen2022templates} on Unseen O-LM, showing the strong generalizability of our methods. \todoo{Our diffusion feature aggregation also achieves better results under the BOP setting. For example, the single-hypothesis average recall increases from 54.7 to 55.5.}
\end{itemize}

A preliminary version of this work is presented as a conference paper~\citep{wang2024object}. In this extended version, we focus on improving the efficiency of the approach by accelerating the feature aggregation process while maintaining the accuracy. To achieve this, we build upon the concept of aggregating intermediate features from diffusion models and introduce the following key improvements: (1) Instead of aggregating all intermediate features indiscriminately, the network now dynamically determines the optimal number of features to aggregate. To enable this, the aggregator is modified to handle a variable number of features, as detailed in Sect.~\ref{subsubsec3-4-1}. (2) We propose a confidence prediction module that computes a confidence score to adaptively  terminate the aggregation process. This enhancement effectively reduces unnecessary computations without compromising the quality of the output, as described in Sect.~\ref{subsubsec3-4-2}. (3) To ensure the efficient training, the loss function is redesigned to supervise the dynamic feature aggregation process. Section~\ref{subsubsec3-4-3} explains how this supervision aligns the aggregation strategy with the model’s overall learning objectives, optimizing both computational efficiency and accuracy. (4) \todoo{We conduct comprehensive experiments to validate the proposed method. Specifically, using the proposed CAA network for feature aggregation reduces computational cost by 1.3-1.5× on the LM, O-LM, and T-LESS datasets compared to our conference version, while maintaining comparable accuracy. In addition, we include new evaluations on the seven core BOP datasets, where integrating diffusion features with CAA \todooo{achieves higher average AR than GigaPose under both single- and five-hypothesis settings.} These results further demonstrate that the proposed approach offers strong generalization to unseen objects and remains effective under the more challenging BOP protocol. Furthermore, we perform extensive analyses and comparative experiments, including evaluations of computational efficiency, feature representation quality, and comparisons with state-of-the-art methods.}

\section{Related Work}
\label{sec:related_work}

Estimating the pose of objects is a fundamental task in computer vision. 
In recent years,  object pose estimation has been dominated by learning-based approaches which achieve great progress. 
\todoo{Thus, in this section, we first review  those commonly employed learning-based approaches.
Since these methods heavily rely on feature extraction and multi-scale representation, feature representation itself also plays a crucial role in accurate pose estimation. Therefore, we further review recent works that adopt U-Net-based architectures and explore feature aggregation mechanisms.}

\paragraph{Indirect methods.} Many learning-based approaches are based on establishing 2D-3D correspondences~\citep{rad2017bb8, park2019pix2pose, li2019cdpn, zakharov2019dpod, hodan2020epos, haugaard2022surfemb}, followed by PnP and RANSAC~\citep{hartley2003multiple, lepetit2009ep} to estimate the pose. These methods primarily focus on the ways of obtaining accurate 2D-3D correspondences. BB8~\citep{rad2017bb8} regresses the 2D coordinates of projected 3D bounding box corners. PVNet~\citep{peng2019pvnet} regresses pixel-wise unit vectors pointing to 2D projections of a set of 3D keypoints. Methods~\citep{park2019pix2pose,li2019cdpn,zakharov2019dpod} employ an encoder-decoder network to regress pixel-level dense correspondences, which are the 2D coordinates of the object surface. 

\paragraph{Direct methods.} Direct methods~\citep{xiang2017posecnn, manhardt2019explaining, do2019real} treat pose estimation as a regression task, directly outputting the object's pose. In addition, SSD-6D~\citep{kehl2017ssd} divides the pose space into categories, transforming it into a classification problem. Some recent methods~\citep{wang2021gdr, chen2022epro}  make the indirect method's PnP process differentiable and use the 2D-3D correspondence from the indirect approach as a surrogate task.

\paragraph{Template-based methods.} Template-based methods
involve determining the object's pose by matching a query image of the object with one of a set of templates, which are rendered images of the 3D model in various poses. One line of methods focuses on obtaining discriminative representations. Method~\citep{wohlhart2015learning}  trains a network to acquire discriminative features of objects and use them for matching during testing. Building on this, method~\citep{balntas2017pose} considers the exact pose differences between training samples to obtain more discriminative representations. AAE~\citep{sundermeyer2018implicit} proposes learning representations by using a denoising autoencoder to recover the original image from an augmented image. Another line of methods includes DeepIM~\citep{li2018deepim}, which proposes iterative refinement by regressing a pose difference between a render of the pose hypothesis and the input image. CosyPose~\citep{labbe2020cosypose} is built on DeepIM, introducing  improvements like a better  architecture, a continuous rotation parametrization, etc.


\paragraph{Unseen object pose estimation.} Recently, unseen object pose estimation has attracted growing interest. Some prior works~\citep{balntas2017pose, li2018deepim, wohlhart2015learning} demonstrated that template-based methods could exhibit a certain degree of generalization to unseen objects. This observation has motivated recent studies~\citep{sundermeyer2020multi, nguyen2022templates, zhao2022fusing} to develop template-based approaches specifically tailored for unseen objects. Due to the inherent difficulty of estimating the pose of objects never observed during training, these methods usually assume that the object is already localized in 2D and focus on predicting its 3D orientation. Although strong performance has been achieved, a noticeable gap between seen and unseen objects still remains. \todoo{
More recent approaches such as MegaPose~\citep{labbe2022megapose} demonstrate that large-scale synthetic training can improve generalization to unseen objects. GigaPose~\citep{nguyen2024gigapose} extends this idea with a very large DINO-based encoder and hundreds of millions of trainable parameters, achieving strong performance at substantially higher computational cost. GenFlow~\citep{moon2024genflow} refines poses by predicting optical flow between rendered and observed images. FoundPose~\citep{ornek2024foundpose}, as a training-free method, relies entirely on frozen pretrained features for pose estimation and the absence of task-specific learning ultimately limits its overall performance. These trends reveal a gap between generalization and efficiency, motivating feature representations that retain strong generalization while requiring only minimal adaptation.
}

\paragraph{Diffusion-based feature aggregation.} \todoo{Recently, several studies explore reusing pretrained diffusion U-Nets (e.g., Stable Diffusion~\citep{rombach2022high}) for dense prediction tasks such as segmentation, detection, and image reconstruction. These works leverage the strong representational ability of diffusion U-Nets by extracting and aggregating multi-scale intermediate features or attention maps from frozen models.
Recent research demonstrates this idea in different contexts. {Diffusion Hyperfeatures}~\citep{luo2023diffusion} applies aggregated diffusion activations to downstream segmentation and classification,
{Diffuse, Attend, and Segment}~\citep{tian2024diffuse} focuses on zero-shot image segmentation using multi-resolution attention, and {Transferring Stable Diffusion to Dense Prediction}~\citep{zhang2024three} investigates how diffusion U-Net features can be adapted to supervised dense prediction tasks. \todooo{
Different from these diffusion-feature extraction methods, {RayPose}
~\citep{huang2025raypose} directly employs a diffusion-based formulation for
pose prediction rather than using a pretrained diffusion model only as a
feature extractor.
} Building upon this trend, our work aggregates intermediate representations from a pretrained diffusion U-Net for object pose estimation. It further introduces a confidence-adaptive early-exit mechanism that improves efficiency while maintaining accuracy, without fine-tuning the diffusion model.}

\section{Methodology}
\label{sec3}

In this section, we first formulate the task of object pose estimation in Sect.~\ref{subsec3-1}; 
We then explore the significance of features and identify the suitability of the intermediate features generated by the text-to-image diffusion model for object pose estimation in Sect.~\ref{subsec3-2}. Next, we propose to  aggregate intermediate features from the diffusion model  for object pose estimation. In this part, we validate the effectiveness of the intermediate features by introducing three distinct networks, as detailed in Sect.~\ref{subsec3-3}. Finally, we propose an  confidence adaptive aggregation network with an early-exiting mechanism to improve efficiency in Sect.~\ref{subsec3-4}.

\subsection{Unseen object pose estimation}\label{subsec3-1}

\begin{figure*}[htb]
\centering
\footnotesize
\rotatebox{90}{\quad \    Template     \quad\quad \ \       Query}
\subfloat{
    \begin{minipage}[t]{0.115\linewidth}
        \centering
        \includegraphics[width=0.75in]{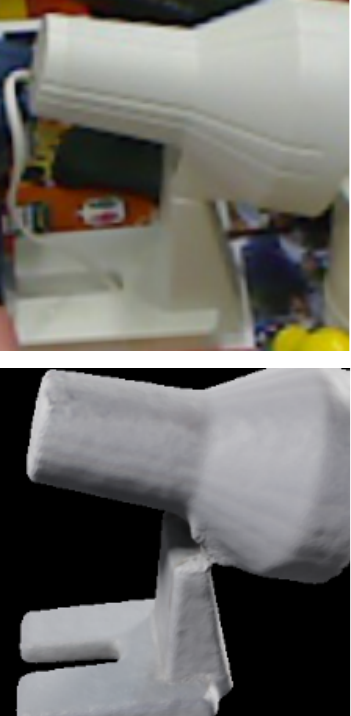}\\
        \centerline{Image}
    \end{minipage}%
        \begin{minipage}[t]{0.115\linewidth}
        \centering
        \includegraphics[width=0.75in]{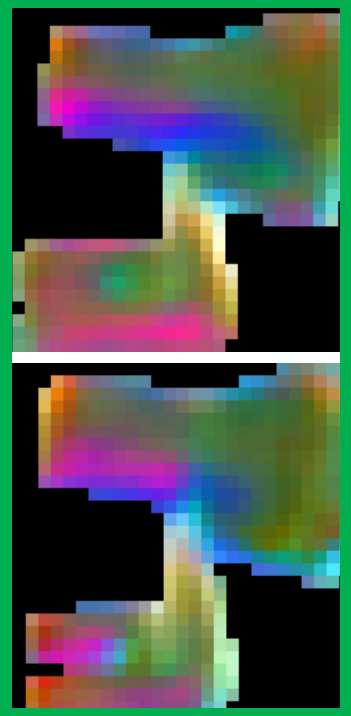}\\
        \centerline{Template-pose}
    \end{minipage}%
        \begin{minipage}[t]{0.115\linewidth}
        \centering
        \includegraphics[width=0.75in]{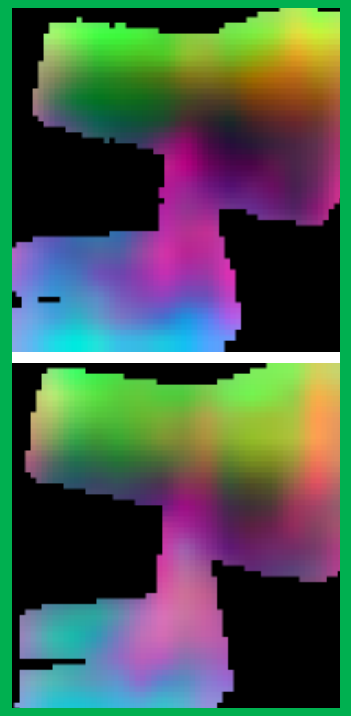}\\
        \centerline{Layer 5}
    \end{minipage}%
        \begin{minipage}[t]{0.115\linewidth}
        \centering
        \includegraphics[width=0.75in]{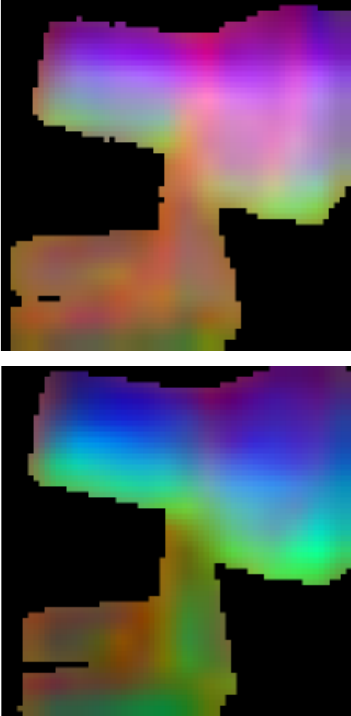}\\
        \centerline{Layer 12}
    \end{minipage}%
}%
\hspace{0.03in}
\subfloat{
    \begin{minipage}[t]{0.115\linewidth}
        \centering
        \includegraphics[width=0.75in]{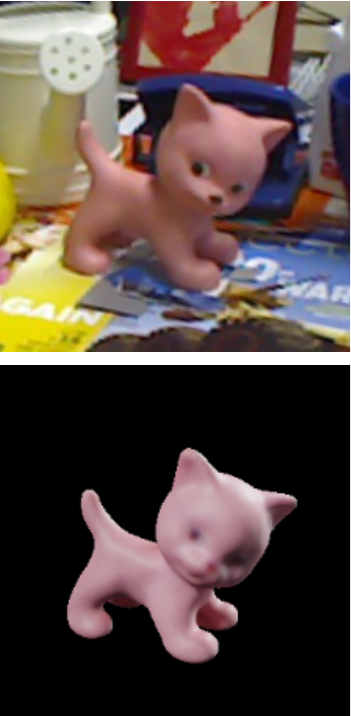}\\
        \centerline{Image}
    \end{minipage}%
    \begin{minipage}[t]{0.115\linewidth}
        \centering
        \includegraphics[width=0.75in]{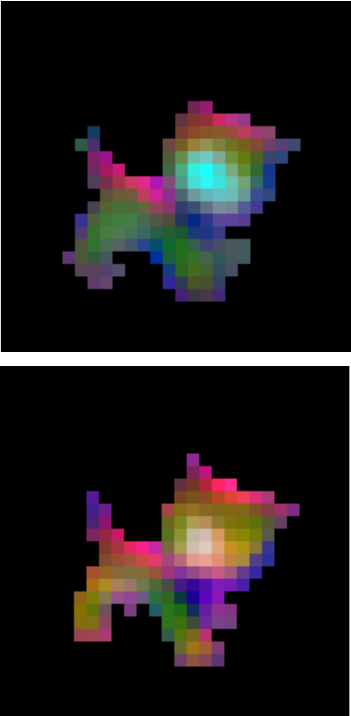}\\
        \centerline{Template-pose
        }
    \end{minipage}%
        \begin{minipage}[t]{0.115\linewidth}
        \centering
        \includegraphics[width=0.75in]{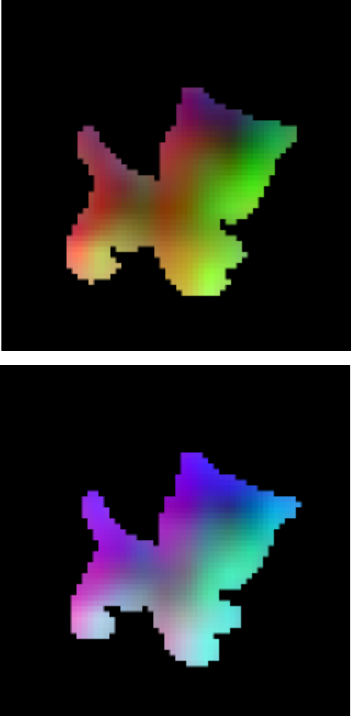}\\
        \centerline{Layer 5}
    \end{minipage}%
        \begin{minipage}[t]{0.115\linewidth}
        \centering
        \includegraphics[width=0.75in]{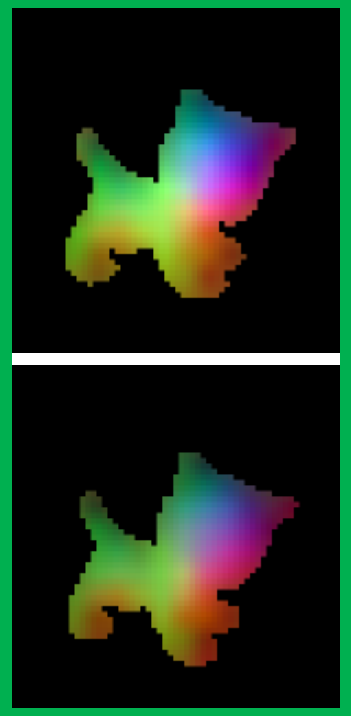}\\
        \centerline{Layer 12}
    \end{minipage}%
}%
\centering
\caption{Feature visualization of LINEMOD. 
For query and template images, we visualize their 3 features from template-pose~\citep{nguyen2022templates}, Layer 5 and Layer 12 of a diffusion model. These features are projected to a PCA space, and the values of top 3 principal components are assigned to RGB values respectively for visualization. The more similar the colors of two feature images in one column, the more similar in feature space. The two features in one green box are very similar.}
\label{fig:feat}
\end{figure*}

\todoo{Given an input object image, the goal of object pose estimation is to determine the rigid transformation from the object to the camera coordinate system, while some methods also infer the object class. Among existing approaches, we focus on template-based methods~\citep{wohlhart2015learning,balntas2017pose,sundermeyer2018implicit,li2018deepim,labbe2020cosypose,sundermeyer2020multi,nguyen2022templates,zhao2022fusing,labbe2022megapose,moon2024genflow,nguyen2024gigapose,ornek2024foundpose} due to their simplicity and promising generalizability. 
These methods estimate object poses by matching an input image against a set of rendered templates that depict the object under different orientations. 
Each template is generated from a 3D model under specific rotation and translation settings, with dense orientation sampling to ensure sufficient coverage of the pose space. 
Depending on the desired level of realism, rendering can range from simple rasterization-based pipelines (e.g., OpenGL) to physically based rendering or ray tracing engines (e.g., BlenderProc~\citep{Denninger2023}), which better simulate illumination, reflection, and material effects. 
Because they rely entirely on synthetic templates rather than object-specific training, template-based methods naturally exhibit strong generalizability and can be applied even to objects not seen during training. 
An \emph{unseen object} in this context refers to an object instance or category absent from the training set, for which the model must estimate the pose solely based on learned visual priors and appearance similarities.}
Given the rendered template set, the input image $I$ and these template images are then mapped into a shared feature space via the encoder $\Phi_{\text{encoder}}$ to compute their similarity:
\begin{equation}
    \mathcal{F} = \Phi_{\text{encoder}}\left(I \right).
\label{eq:feat}
\end{equation}
The pose $\hat{P}$ of the most similar template image is then assigned to the given real image to achieve the final pose estimation.
In some cases, the object class $\hat{c}$ is inferred together with the pose.

Clearly, acquiring an effective feature $\mathcal{F}$ through Eq.~(\ref{eq:feat}) is pivotal for computing the similarity between real and template images, and therefore is crucial for successful object pose estimation.

\subsection{Motivation: \todoo{features matter}}\label{subsec3-2}

In recent years, object pose estimation has achieved significant success, e.g. template-based methods~\citep{wohlhart2015learning, balntas2017pose, sundermeyer2020multi}, on seen objects. However, these methods tend to struggle when dealing with \emph{unseen} objects. To address this issue, recent methods~\citep{nguyen2022templates,zhao2022fusing} propose to  encode images into local features, resulting in interesting performance for unseen objects. In spite of the considerable progress, their methods still exhibit a noticeable performance gap between seen and unseen objects.

Despite the advancements in feature engineering, generalizing these improvements to  \emph{unseen}  objects remains a significant challenge. 
Figure \ref{fig:feat} illustrates this challenge by selecting two images (query) from the LM dataset. One image contains a \emph{seen} object, the {lamp}, while the other showcases an \emph{unseen} {cat}. Subsequently, we render the corresponding template images (e.g. the leftmost one in the 2nd row) using the same  pose as the query. 
We then visualize the features of query and template in the feature space of the state-of-the-art method~\citep{nguyen2022templates}. We project these features to a PCA space and assign RGB channels for visualization. 
The more similar color channels indicate more similar features. Remarkably, for the \emph{seen} object lamp, the  features of  the query and template exhibit significant similarity. However, for  the \emph{unseen} object cat, notable differences emerge. This discovery underscores the primary focus of this paper: the pursuit of a generalized object features to accurately estimate the pose of  unseen objects.

To obtain the discriminative features with strong generalizability, we realize the recent text-to-image diffusion models  have  demonstrated strong features to generate high-quality images. {As discussed in Sect.~\ref{sec:intro}, a number of recent studies~\citep{zhan2023does,xu2023open,tang2023emergent,luo2023diffusion,li2023guiding} have demonstrated that diffusion features generalize well to different scenarios.}

Since image features are important for object pose estimation and the  diffusion features can potentially offer discriminative features, we are inspired to investigate the feasibility of using diffusion features for object pose estimation. To verify this assumption, we employ a well-known diffusion model, Stable Diffusion~\citep{rombach2022high},  for this investigation.
When an image is fed into Stable Diffusion, it generates a set of intermediate features from a UNet. 
As illustrated in Fig. \ref{fig:feat}, we visualize the intermediate features from both Layer 5 and Layer 12 of the UNet  at timestep $t=0$. 
Interestingly, the features from Layer 5 can well measure the similarity of query and template for the object lamp; in contrast, Layer 12 is the best for the object cat. 

\paragraph{Summarization:} 
(1) The features of the state-of-the-art method~\citep{nguyen2022templates} cannot well measure the similarity of unseen objects; (2) The features of Stable Diffusion show great potential to model \emph{unseen} objects. For example, we do not use the lamp and cat images from LINEMOD dataset to finetune the model, however, Stable Diffusion performs quite well;  (3) The features of Stable Diffusion from different layers can capture different dynamics, e.g. Layer 5 and 12 are the best for the lamp and cat  respectively. It inspires us to investigate the aggregation of features from different layers to achieve strong generalizability on unseen objects.

\subsection{Exploring diffusion feature aggregation}\label{subsec3-3}

In this section, we first introduce the diffusion process in Sect.~\ref{subsubsce:diffusion process}. We then explore various approaches for aggregating diffusion features in  Sect.~\ref{subsubsce:diffusion feature aggregetion}. Finally, we  use the aggregated features from the diffusion model for object pose estimation in Sect.~\ref{subsubsce:traintest}.

\subsubsection{Diffusion process}
\label{subsubsce:diffusion process}

Text-to-image diffusion models involve both forward and reverse processes. One widely used sampling procedure for diffusion models is DDIM~\citep{song2020denoising}, as defined:
\begin{equation}
x_t = \sqrt{\alpha_t} x_0 + \sqrt{1-\alpha_t} \epsilon_t, \text { where } \epsilon_t \sim \mathcal{N}(0,1),
\end{equation} 
where $x_0$ denotes the initial clean sample, $\alpha_t$ denotes the pre-defined noise schedule, and $\epsilon_t$ denotes the randomly generated noise. During the forward process, Gaussian noise is gradually added to the clean sample over $T$ steps, resulting in a sequence of noisy samples: $x_1, x_2, \ldots, x_T$. As $T$ approaches infinity, $x_T$ converges to an isotropic Gaussian distribution. In the reverse process, the diffusion model is sampled by iteratively denoising $x_T \sim \mathcal{N}(0,1)$ conditioned on a text prompt $y$. Specifically, at each denoising step for $t=1, \ldots, T$, $x_{t-1}$ is determined using both $x_t$ and the text prompt $y$. After the final denoising step, $x_0$ is transformed back to the clean sample.

Popular diffusion-based text-to-image generation models~\citep{ramesh2022hierarchical, rombach2022high, saharia2022photorealistic} commonly use a UNet architecture to learn the denoising process. The UNet consists of convolutional blocks, upsampling and downsampling blocks, skip connections, and attention blocks that apply cross-attention between text embeddings and UNet features. The feature maps from the UNet blocks can be regarded as rich and dense features, which are highly suitable for object pose estimation.

In practice, our objective is to obtain features directly from clean images without relying on conditional prompts. Specifically, we achieve this by 
employing an unconditioned text embedding 
and 
running Stable Diffusion only once with a very small timestep, such as $t=0$. With such a small timestep, from the perspective of the diffusion model, clean images are predominantly treated as denoised  images. 

\subsubsection{Feature aggregation}
\label{subsubsce:diffusion feature aggregetion}

\begin{figure*}[!t]
\centering
\footnotesize
\includegraphics[width=0.95\textwidth]{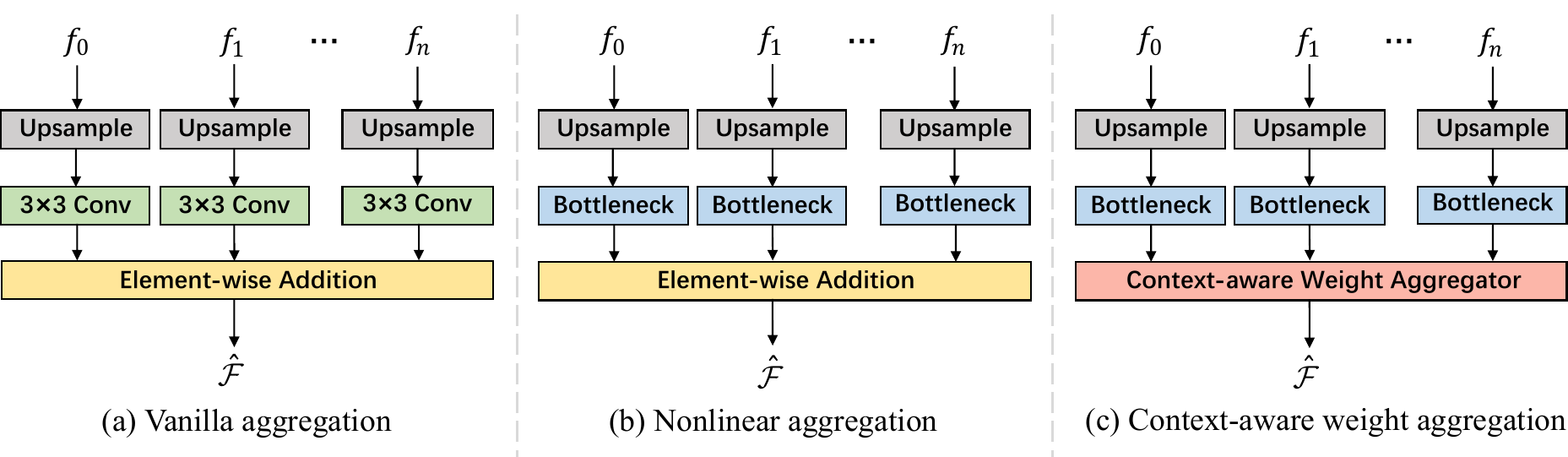}
\caption{Diffusion aggregation methods aggregate all intermediate features.}     
\label{fig:modules-abc}
\end{figure*} 

In Sect.~\ref{subsubsce:diffusion process}, we introduce the feature extraction from a diffusion model to validate the efficacy of diffusion features. Motivated by the findings in Fig. \ref{fig:feat}, the aggregation of features extracted from different layers can potentially generalize well to different objects and scenarios. Thus, in this section, we investigate the ways of aggregation.

Given a set of raw diffusion  features $\{f_1, f_2,  \ldots, f_{n}\}$ from  $n$ layers in a diffusion model, we need to employ  trainable  architectures to aggregate these features into a fused one $\hat{\mathcal{F}}$. The aggregation involves two essential components, an {extractor}  $\Phi_{\text{ext}}$ and an aggregator $\Phi_{\text{aggr}}$. Specifically, $\Phi_{\text{ext}}$ is employed to align the raw diffusion features to the same dimension and to learn task-specific features  adapting to object pose estimation;
while $\Phi_{\text{aggr}}$ is employed to aggregate the extracted features. This process can be standardized as follows: 
\begin{equation}
    \hat{\mathcal{F}} = \Phi_{\text{aggr}} \left( \{\Phi_{\text{ext}}^1\left(f_1 \right), \ldots,\Phi_{\text{ext}}^n\left(f_n \right) \} \right).
\end{equation} 
In this section we discuss several aggregation architectures for diffusion features.

\paragraph{Vanilla  aggregation.} First, we design a Vanilla  Aggregation (VA) network, illustrated in Fig. \ref{fig:modules-abc}(a). Given a set of diffusion  features, we first upsample them to a standard resolution. Next, we use $3\times 3$ convolution layers to learn features specialized for object pose estimation 
and project them to the same channel count. Finally, we aggregate them directly through element-wise addition as: \begin{equation}
\hat{\mathcal{F}} = \sum_{i=1}^n \Phi_{\text{ext}}^i(f_i).
\end{equation} 

\paragraph{Nonlinear aggregation.} Since the vanilla aggregation network only performs linear mapping on the original features and lacks any nonlinearity, it falls short in capturing complex data patterns and nonlinearities. Hence, we design a Nonlinear Aggregation (NA) network, illustrated in Fig. \ref{fig:modules-abc}(b). In this architecture, we substituted $\Phi_{\text{ext}}$ from the vanilla aggregation network with a bottleneck layer~\citep{he2016deep} to introduce nonlinearity. 
Specifically, the bottleneck layer consists of three convolutions and ReLU functions with a skip connection.
However, as the number of parameters increases, the limited training data adversely affects the generalization of the aggregated features. In response to this challenge, we take inspiration from ControlNet~\citep{zhang2023adding} and initialize the last convolution in the bottleneck with zero values.

\paragraph{Context-aware weight aggregation.} 
During aggregation, VA and NA implicitly set the weights of each feature as 1 for element-wise addition. However, the optimal weights can be learned to lead to better aggregation. Thus, we design a so-called Context-aware Weight Aggregation (CWA) network that learns the weights based on the context, as illustrated in Fig. \ref{fig:modules-abc}(c). Specifically, we start by upsampling all features and then passing them through an extractor composed of bottleneck layers same as NA:
\begin{equation}
h_i = \Phi_{\text{ext}}^i(f_i).
\end{equation}
Subsequently, we apply an average pooling layer to transform each $h_i$ into one-dimensional features $z_i$. Following this, we concatenate all these one-dimensional features to represent the entire context. Finally, we employ a multilayer perceptron (MLP) and a softmax to determine the weights associated with each feature:
\begin{equation}
\{\mathbf{w}_1, \mathbf{w}_2, \ldots, \mathbf{w}_n\} = \text{softmax}(\text{MLP}([z_1, z_2, \ldots, z_n])).
\end{equation} 
Finally, these weights are used to perform a weighted sum of all the features to achieve the aggregation:
\begin{equation}
\hat{\mathcal{F}}=\sum_{i=1}^n \mathbf{w}_i \cdot h_i.
\label{eq:final_all}
\end{equation}

\subsubsection{Strategies of training and test}
\label{subsubsce:traintest}

To adapt our method for object pose estimation, we freeze the weights of SD model and only train  our aggregation networks under  the supervision of  pose estimation. 
Following the approach outlined in template-pose~\citep{nguyen2022templates}, we create $1$ positive pair template and $M-1$ negative pair templates for each real image. Our model is trained to maximize the agreement between the representations of samples in positive pairs while minimizing that of negative pairs. We achieve this using the InfoNCE~\citep{oord2018representation} loss $\mathcal{L}_{NCE}(\hat{\mathcal{F}}^{\text{real}}, \hat{\mathcal{F}}^{\text{tmpl}})$, where $\hat{\mathcal{F}}^{\text{real}}$ represents the feature representation of the real image and $\hat{\mathcal{F}}^{\text{tmpl}}$ represents the feature representation of the template image.

To estimate the object pose in the input image, we first identify the template that closely resembles the input. We then transfer the identity and pose information annotated in the template to the input image, completing the pose estimation process.



\subsection{Confidence adaptive aggregation}\label{subsec3-4}

\begin{figure*}[!t]
\centering
\footnotesize
\includegraphics[width=0.95\textwidth]{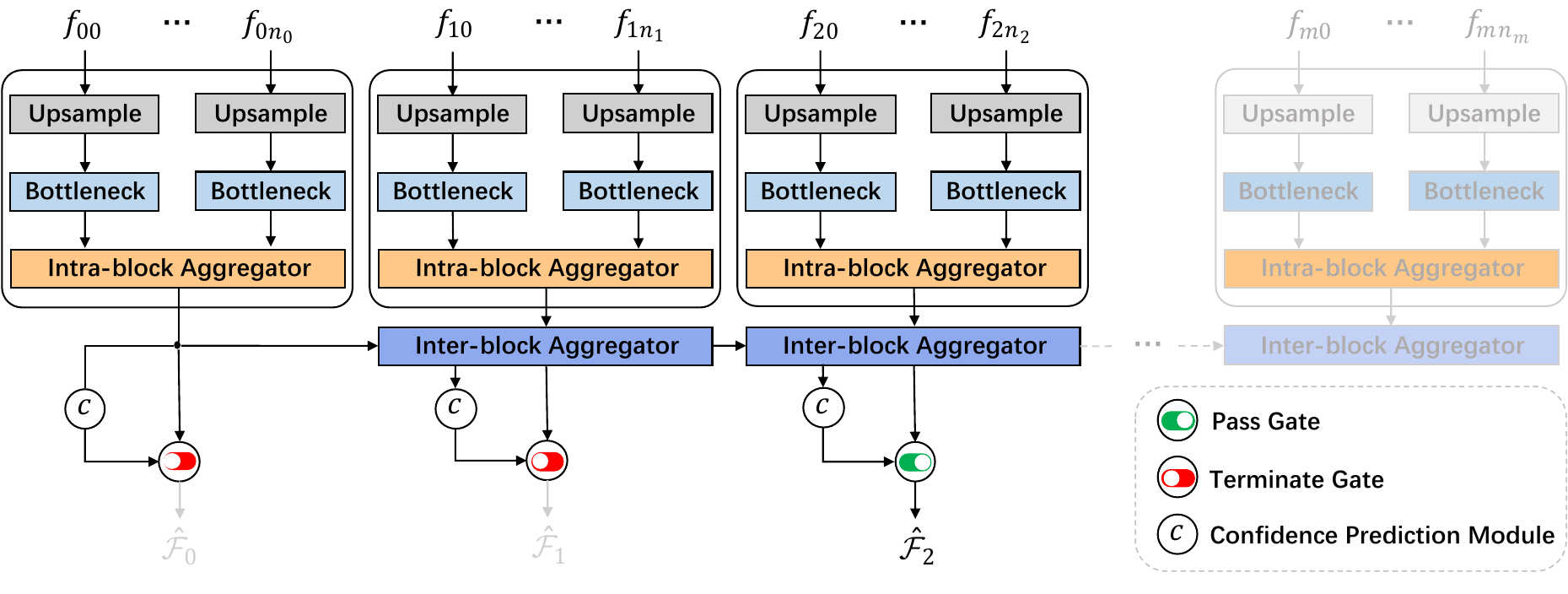}
\caption{Confidence Adaptive Aggregation (CAA) method aggregates features within and across blocks, with early exit decisions guided by confidence prediction modules.} 
\label{fig:modules-d}
\end{figure*} 

In real applications, it is essential not only to generalize to unseen objects but also to achieve real-time performance. 
Though the methods in Sect. 3.3 can achieve very promising accuracy, in some applications, we need a  better speed-and-accuracy tradeoff. 
Motivated by this, we introduce a very efficient network, Confidence Adaptive Aggregation (CAA), as illustrated in Fig. \ref{fig:modules-d}. This network selectively aggregates features and  makes early exit decisions based on confidence scores, significantly reducing computational costs without sacrificing accuracy.

Confidence Adaptive Aggregation (CAA) learns  to aggregate the
features whose  confidence  is sufficiently high, thereby avoiding   aggregating all intermediate features $\{f_1, f_2, \ldots, f_{n}\}$ to achieve an efficient inference. Specifically, at any intermediate layer $i < n$, the confidence score $C_i$ is calculated based on the previously aggregated features ${f_1, f_2, \ldots, f_{i}}$. To accomplish this, three key components are necessary for the aggregation network: 1) developing an aggregation method that can handle variable numbers of features, as discussed in Sect.~\ref{subsubsec3-4-1}; 2) designing a module that accurately predicts the confidence score, elaborated in Sect.~\ref{subsubsec3-4-2}; and 3) training the network to predict confidence scores while enabling early exits without compromising accuracy, detailed in Sect.~\ref{subsubsec3-4-3}. Finally, in Sect.~\ref{subsubsec3-4-4}, we use   the confidence score   to guide early exits for efficient object pose estimation.

\subsubsection{Dynamic feature aggregator}\label{subsubsec3-4-1}

The previous aggregation network always aggregates all intermediate features together. To handle variable numbers of features, we propose a dynamic feature aggregator.
In Stable Diffusion, the UNet model consists of $m$ blocks, with each block $B_j$ containing $l_j$ layers, where $j\in \{1, 2, \ldots, m\}$.
We aggregate the features within each block and its preceding blocks into a single feature. 
For the intermediate features of each layer within each block, we use ResNet bottleneck as extractor $\Phi_{\text{ext}}^{j,k}$, similar to previous methods:
\begin{equation}
h_{j,k} = \Phi_{\text{ext}}^{j,k}(f_{j,k}),
\end{equation}
where $f_{j,k}$ represents the intermediate feature of the $k$-th layer in the $j$-th block $B_j$.
Additionally, unlike previous methods, we employ different aggregators for intra- and inter-block features to achieve more effective feature representations{, as shown in Fig. \ref{fig:modules-d}.}
Specifically, in each block $B_j$, we use average pooling as the intra-block aggregator $\Phi_{\text{intra}}$:
\begin{equation}
\Phi_{\text{intra}}(B_j) = \frac{1}{l_j} \sum_{k=1}^{l_j} \Phi_{\text{ext}}^{j,k}(f_{j,k}).
\end{equation}
While between blocks, we use max pooling as the inter-block aggregator $\Phi_{\text{inte}}$:
\begin{equation}
\begin{split}
\Phi_{\text{inter}}&= \text{maxpool}\{\Phi_{\text{intra}}(B_1), \Phi_{\text{intra}}(B_2), \\
&\quad \ldots, \Phi_{\text{intra}}(B_j)\},
\end{split}
\end{equation}
We aggregate the features of block $B_j$ and the preceding blocks to obtain the aggregated feature $\hat{\mathcal{F}_j}$:
\begin{equation}
\hat{\mathcal{F}_j} = \Phi_{\text{inter}}\left( \{\Phi_{\text{intra}}(B_q) \mid q = 1, 2, \ldots, j\} \right).
\end{equation}

\subsubsection{Confidence prediction module}\label{subsubsec3-4-2}

In addition to dynamically aggregating intermediate features, the network must predict a confidence score to determine when to exit. We design a confidence prediction module, $\Psi$, to obtain a confidence score $C_j$ from each aggregated feature $\hat{\mathcal{F}_j}$, as shown in Fig. \ref{fig:modules-d}.
Specifically, we first apply two 3x3 convolutions with ReLU activation functions and average pooling to reduce the feature dimensions. This is followed by a fully connected layer with a Tanh activation function to estimate the confidence score:
\begin{equation}
C_j = \text{Tanh}(\Psi(\hat{\mathcal{F}_j})).
\end{equation}
The Tanh activation function restricts the output to the range $(-1, 1)$. 
In our experiments, we share the confidence prediction module across all blocks to improve the training efficiency and prediction consistency of the entire model.

\subsubsection{Joint training for confidence and feature aggregation}\label{subsubsec3-4-3}
As mentioned in Sect.~\ref{subsec3-1}, 
\todooo{unseen object pose estimation is performed by retrieving the template that is most similar to the input image. Accordingly, rather than directly supervising the confidence score with binary retrieval correctness, we use feature similarity to the corresponding positive template as its supervision. Therefore, the predicted confidence should be interpreted as a similarity-based retrieval-quality or stopping score, rather than as a directly supervised probability of correct retrieval. A higher confidence score indicates that the progressively aggregated query representation is more similar to the corresponding positive template representation, suggesting that the current representation is more suitable for reliable template retrieval.}

\todooo{To construct this similarity-based supervision at each aggregation block, we
compare the progressively aggregated representation of the real image with a
reference representation of its corresponding positive template.} Specifically, for the template image, 
we aggregate all intermediate features of UNet in Stable Diffusion to obtain the aggregated feature $\hat{\mathcal{F}}^{\text{tmpl}}$.
For the real image, we progressively aggregate intermediate features from the first block up to block $j$, yielding a comprehensive feature representation, $\hat{\mathcal{F}}_j^{\text{real}}$.
We calculate the cosine similarity between the two features as the supervision for the confidence score:
\begin{equation}
\text{sim}(\hat{\mathcal{F}}_j^{\text{real}}, \hat{\mathcal{F}}^{\text{tmpl}}) = \frac{\hat{\mathcal{F}}_j^{\text{real}} \cdot \hat{\mathcal{F}}^{\text{tmpl}}}{\lVert \hat{\mathcal{F}}_j^{\text{real}}\lVert \lVert \hat{\mathcal{F}}^{\text{tmpl}} \lVert}.
\end{equation}
The loss function is designed as the MSE loss of the estimated confidence score $C_j$ and the feature similarity ground truth:
\begin{equation}
\mathcal{L}_{\text{conf}} = \sum_{j=1}^m\lVert C_j - \text{sim}(\hat{\mathcal{F}}_j^{\text{real}}, \hat{\mathcal{F}}^{\text{tmpl}}) \rVert ^2.
\end{equation}
To encourage the model to perform early exits while maintaining accuracy close to that of the full mode, we propose updating each block simultaneously using the pose estimation loss.
Specifically, we apply the InfoNCE loss to all aggregated features and average them:
\begin{equation}
\mathcal{L}_{\text{avgNCE}} = \frac{1}{m} \sum_{j=1}^m \mathcal{L}_{\text{NCE}}(\hat{\mathcal{F}}_j^{\text{real}}, \hat{\mathcal{F}}^{\text{tmpl}}).
\end{equation}
We use a joint training strategy to simultaneously optimize the InfoNCE loss and the confidence loss:
\begin{equation}
\mathcal{L}_{\text{all}} = \mathcal{L}_{\text{avgNCE}} + \lambda \mathcal{L}_{\text{conf}},
\end{equation}
where $\lambda$ is a hyperparameter used to balance the two losses. In our experiments, we set $\lambda$ to 1.

\subsubsection{Object pose estimation with early-exiting}\label{subsubsec3-4-4}

After training, our network not only predicts confidence scores but can also perform early exits at any intermediate block. For each image requiring pose estimation, we first feed it into Stable Diffusion, processing one block of the U-Net at a time. At each block, our dynamic feature aggregator aggregates all the intermediate features, and the aggregated features are then passed to a confidence prediction network. 
As illustrated in Fig. \ref{fig:modules-d}, the aggregated features from each block are evaluated with a confidence score, enabling the network to determine if it should exit early based on a predefined threshold. In our experiments, this threshold, $\delta$, is set to 0.7.
If the confidence score exceeds this threshold, the network terminates the process and uses the aggregated feature from that block as the final output. The pose estimation is then carried out using the method outlined in Sect.~\ref{subsubsce:traintest}.

\section{Experiments}
\label{sec:exp}

In this section, we first introduce the implementation details, including the datasets and evaluation metrics used (Sect.~\ref{sub:setup}). We then compare our method with state-of-the-art approaches on the LM, O-LM, and T-LESS datasets (Sect.~\ref{sub:com}). This is followed by an ablation study that investigates the effects of key design choices on performance, including timestep selection, the impact of models pre-trained at large scale, confidence threshold settings, and other related factors (Sect.~\ref{sub:ablation}).
To further illustrate the effectiveness of our method, we present visualizations and qualitative analyses (Sect.~\ref{sub:quali_any}). Finally, we compare the number of trainable parameters across different methods (Sect.~\ref{sub:efficiency}).

\subsection{Experimental  settings}
\label{sub:setup}
\subsubsection{Implementation details}
\label{sub:imp}

In our experiments, we extract features from Stable Diffusion (SD) v1-5~\citep{rombach2022high}, a generative latent diffusion model trained on LAION-5B~\citep{schuhmann2022laion}. 
We feed images with a resolution of 512 into this SD model and aggregate features from all layers of its UNet into output features with a dimensionality of 32$\times$32. We train our network for 20 epochs with a learning rate of 1e-3 for LM dataset and 1e-4 for T-LESS dataset. Throughout all experiments, we directly feed images  into the SD model without adding any noise. For the Confidence Adaptive Aggregation, we set the confidence threshold $\delta$ to 0.7 to achieve a trade-off between accuracy and computational costs.


\subsubsection{Dataset}
\label{subsub:dataset}

Following existing template-based works such as Template-Pose~\citep{nguyen2022templates}, we evaluate our method on three widely used benchmarks, namely LINEMOD (LM)~\citep{hinterstoisser2013model}, Occlusion-LINEMOD (O-LM)~\citep{brachmann2014learning}, and T-LESS~\citep{hodan2017t}, which represent the standard setting for comparisons within template-based pose estimation. \todoo{To further assess generalization in a broader \emph{unseen} object setting, we also evaluate our method under the BOP Challenge~\citep{sundermeyer2023bop}, using the large-scale synthetic training data provided by MegaPose~\citep{labbe2022megapose}.
}

\paragraph{LM and  O-LM.} The LM~\citep{hinterstoisser2013model} dataset consists of 13 sequences, each containing ground truth poses for a single object. CAD models for all the objects are provided. The O-LM~\citep{brachmann2014learning} dataset is an extension of the LM dataset, with objects in the dataset being heavily occluded, making pose estimation highly challenging. 
We follow the approach outlined in Template-Pose~\citep{nguyen2022templates} to divide the cropped data into three non-overlapping groups based on the depicted objects. For each split, the unseen objects are defined as follows:
\begin{itemize}
    \item Split \#1: \textit{Ape}, Benchvise, Camera, \textit{Can};
    \item Split \#2: \textit{Cat}, \textit{Driller}, \textit{Duck}, \textit{Eggbox};
    \item Split \#3: \textit{Glue}, \textit{Holepuncher}, Iron, Lamp, Phone.
\end{itemize}
In each split, the remaining objects from the LM dataset are used for training. Objects listed in \textit{italics} are frequently occluded in the O-LM dataset. Additionally, we reserve 10\% of the data for testing on seen objects. Our model is trained on LM and evaluated on both LM and O-LM.
Following the protocol of Wohlhart \textit{et al.}~\citep{wohlhart2015learning}, we render 301 templates per object. The input image is cropped at the center of objects with the ground-truth pose, following the approach used in previous works~\citep{wohlhart2015learning, balntas2017pose, nguyen2022templates}.



\paragraph{T-LESS.} T-LESS~\citep{hodan2017t} comprises 30 objects characterized by a lack of distinct textures or distinguishable colors. These objects are symmetric and similar in shape and size, presenting challenging scenarios with substantial clutter and occlusion. For T-LESS, we adopt the testing methodology established in previous works~\citep{nguyen2022templates, sundermeyer2020multi}. We split the 30 objects into two groups: objects 1-18 are considered as seen objects, and objects 19-30 as unseen objects. Our model is trained on the dataset that includes only the seen objects, and testing is performed on the complete test dataset. Following the protocol of template-pose~\citep{nguyen2022templates}, we render $21,672$ templates per object. Similar to previous works~\citep{sundermeyer2020multi, sundermeyer2018implicit, nguyen2022templates}, we use the ground-truth bounding box to crop the input image.

\paragraph{MegaPose.} \todoo{The MegaPose~\citep{labbe2022megapose} training set is a large-scale synthetic dataset constructed for \emph{unseen} object pose estimation. It contains approximately two million rendered RGB images generated with BlenderProc~\cite{Denninger2023}, using objects sampled from ShapeNet (SN)~\citep{chang2015shapenet} and Google-Scanned Objects (GSO)~\citep{downs2022google}. During synthesis, object placement, materials, illumination, backgrounds, and camera viewpoints are extensively randomized. MegaPose also adopts training configurations that combine approximately $1000$ GSO objects with up to $20000$ ShapeNet objects, providing broad geometric and visual diversity.}

\paragraph{BOP challenge.} \todoo{The BOP challenge~\citep{sundermeyer2023bop} provides a comprehensive \emph{unseen} object evaluation setting across seven core datasets: LineMod-Occlusion (LM-O)~\citep{brachmann2014learning}, T-LESS~\citep{hodan2017t}, TUD-L~\citep{sundermeyer2023bop}, ICBIN~\citep{doumanoglou2016recovering}, ITODD~\citep{drost2017introducing}, HomebrewedDB (HB)~\citep{kaskman2019homebreweddb}, and YCB-Video (YCB-V)~\citep{xiang2017posecnn}.
Across these datasets, 132 object models appear exclusively in the test scenes, forming a challenging benchmark for evaluating generalization to novel objects. The test split contains 19,048 annotated instances captured in cluttered and occluded environments, providing a demanding real-world setting for assessing robustness.}

\subsubsection{Evalutation metrics}
\label{subsub:metric}

For the LM and O-LM datasets, we calculate pose error by computing the geodesic distance~\citep{wohlhart2015learning} between two rotations:
\begin{equation}
\todooo{
d(\hat{\mathbf{R}},\mathbf{R})
=
\arccos\left(
\frac{\operatorname{tr}(\mathbf{R}^{T}\hat{\mathbf{R}})-1}{2}
\right)
\frac{180}{\pi}.
}
\end{equation}
Here, $\mathbf{R}$ denotes the ground truth 3D orientation, $\hat{\mathbf{R}}$ denotes the predicted 3D orientation, and $\text{tr}$ denotes the trace. In the case of an unseen object, the object class is unknown during testing. The accuracy is defined as the percentage of test images for which the best angle error is below a specific threshold and the predicted object class is correct. It can be computed as follows:
\begin{equation}
\mathrm{Acc.} =
\begin{cases}
1, & \text{if } d(\hat{\mathbf{R}},\mathbf{R}) < \lambda
     \text{ and } \hat{c}=c,\\
0, & \text{otherwise}.
\end{cases}
\end{equation}
Here, $c$ denotes the ground truth class. Following template-pose~\citep{nguyen2022templates}, we use the Acc15 metric, where the angular threshold $\lambda$ is set to $15^\circ$.

Since most of the objects in the T-LESS dataset are symmetric, we follow the evaluation protocol established in previous works~\citep{hodavn2016evaluation, sundermeyer2018implicit, sundermeyer2020multi, nguyen2022templates}. Specifically, we report the recall under the Visible Surface Discrepancy (VSD) metric with a threshold of 0.3. Additionally, we estimate the translation of the object using the projective distance estimation method from SSD-6D~\citep{kehl2017ssd}, which has been widely adopted in previous studies~\citep{sundermeyer2018implicit, sundermeyer2020multi, nguyen2022templates}. This method calculates the translation by analyzing the 3D transformation between the object template and the input bounding box, enabling accurate 3D position prediction.

To evaluate the computational expense of the methods, we measure the MACs (Multiply-Accumulate Operations) and \todoo{FPS (Frames Per Second)} for each method. MACs quantify the total number of multiplication and addition operations performed by the method, providing insight into its computational complexity. FPS measures the rate at which the method processes data, reflecting its efficiency in handling large volumes of information. We measure FPS on an A6000 GPU to ensure consistency and reliability in our comparisons. By analyzing both MACs and FPS, we aim to assess the trade-offs between computational demands and processing speed, thereby gaining a better understanding of the method's overall accuracy and efficiency.

\todoo{We evaluate our method on the BOP benchmark following the official BOP Challenge protocol~\citep{hodavn2020bop}. The evaluation uses three complementary metrics. Visible Surface Discrepancy (VSD) compares only the visible object regions and naturally accounts for pose ambiguities. Maximum Symmetry-Aware Surface Distance (MSSD) measures 3D surface deviation using predefined global symmetries. Maximum Symmetry-Aware Projection Distance (MSPD) quantifies perceptual deviation in the image plane while respecting the same symmetries. A pose prediction is considered correct under each metric when its error falls below the corresponding threshold, and the proportion of correct predictions defines the Recall. The Average Recall for each metric is obtained by integrating over a range of thresholds, and for VSD, over multiple visibility tolerance settings as well. The final performance score, Average Recall (AR), is computed as the mean of the three metric-specific values.}

\subsection{Comparison with the state-of-the-art}

\label{sub:com}


\subsubsection{LINEMOD and Occluded-LINEMOD}\label{sub:com-lm}

\begin{table*}[htb]
\centering
\caption{\todoo{Results on LM~\citep{hinterstoisser2013model}. VA represents vanilla aggregation, NA represents nonlinear aggregation, CWA represents context-aware weight aggregation, and CAA represents confidence adaptive aggregation.}}
\resizebox{\linewidth}{!}
{\begin{tabular}{l|l|c|cc|cc|cc|cccc}
\hline
\multirow{2}{*}{\textbf{Method}} & \multirow{2}{*}{\textbf{Backbone}} & \multirow{2}{*}{\textbf{Res.}} & \multicolumn{2}{c|}{\textbf{Split \#1}} & \multicolumn{2}{c|}{\textbf{Split \#2}} & \multicolumn{2}{c|}{\textbf{Split \#3}} & \multicolumn{4}{c}{\textbf{Avg.}} \\
 & & & {Seen} & \textbf{Unseen} & {Seen} & \textbf{Unseen} & {Seen} & \textbf{Unseen} & {Seen} & \textbf{Unseen} & MACs & FPS \\ \hline
Balntas \textit{et al.}~\citep{balntas2017pose} & Wohlhart \textit{et al.}~\citep{wohlhart2015learning} & $64^2$ & 87.0 & 13.2 & 83.1 & 15.5 & 85.1 & 18.2 & 85.0 & 15.2 & 11.9M & 493 \\
Wohlhart \textit{et al.}~\citep{wohlhart2015learning} & Wohlhart \textit{et al.}~\citep{wohlhart2015learning} & $64^2$ & 89.2 & 14.1 & 85.4 & 16.3 & 83.3 & 16.7 & 86.3 & 16.7 & 11.9M & 493 \\
MPL~\citep{sundermeyer2020multi} & MPL~\citep{sundermeyer2020multi} & $128^2$ & 91.5 & 36.0 & 87.9 & 39.3 & 87.2 & 39.7 & 88.9 & 38.3 & 2.1B & 52.4 \\
Zhao \textit{et al.}~\citep{zhao2022fusing} & ResNet~\citep{he2016deep} & $128^2$ & 96.9 & 87.5 & 94.4 & 76.2 & 93.1 & 73.7 & 93.8 & 79.2 & 1.7B & 43.8 \\ 
Template-pose~\citep{nguyen2022templates} & ResNet~\citep{he2016deep} & $512^2$ & 99.3 & 95.8 & 99.1 & 97.9 & 99.4 & 89.6 & 99.3 & 94.4  & 130B & 46.5 \\ 
Template-pose~\citep{nguyen2022templates} & ResNet~\citep{he2016deep} & $224^2$ & 99.3 & 94.4 & 99.0 & 97.4 & 99.2 & 88.7 & 99.1 & 93.5 & 24.9B & 136 \\ \hline

Ours (VA) & SD-V1-5~\citep{rombach2022high} & $512^2$ & 99.4 & 97.4 & 99.5 & 99.3 & 99.4 & 95.6 & 99.4 & 97.4 & 902B & 10.2 \\
Ours (NA) & SD-V1-5~\citep{rombach2022high} & $512^2$ & 99.5 & 97.5 & 99.6 & \textbf{99.6} & 99.7 & 96.0 & 99.6 & 97.7 & 929B & 9.5 \\
Ours (CWA) & SD-V1-5~\citep{rombach2022high} & $512^2$ & 99.5 & \textbf{98.3} & 99.6 & 98.9 & \textbf{99.8} & \textbf{96.3} & \textbf{99.7} & \textbf{97.9} & 929B & 9.2 \\ 
Ours (CAA) & SD-V1-5~\citep{rombach2022high} & $512^2$ & \textbf{99.7} & 97.4 & \textbf{99.7} & 99.4 & 99.5 & 96.2 & 99.6 & 97.7 & 678B & 15.5 \\
Ours (CAA) & SD-V1-5~\citep{rombach2022high} & $256^2$ & 99.4 & {97.7} & 99.6 & 99.5 & 99.3& 96.1 & 99.4 & 97.8 & 169B & 45.3 \\
Ours (CAA) & SD-V1-5~\citep{rombach2022high} & $128^2$ & 99.2 & 97.1 & 99.4 & 99.1 & 99.3 & 96.0 & 99.3 & 97.4 & 42.3B & 68.1 \\\hline
\end{tabular}}

\label{tab:lm}
\end{table*}

\begin{table*}[htb]
\centering
\caption{\todoo{Results on O-LM~\citep{brachmann2014learning}. VA represents vanilla aggregation, NA represents nonlinear aggregation, CWA represents context-aware weight aggregation, and CAA represents confidence adaptive aggregation.}}
\resizebox{\linewidth}{!}
{\begin{tabular}{l|l|c|cc|cc|cc|cccc}
\hline
\multirow{2}{*}{\textbf{Method}} & \multirow{2}{*}{\textbf{Backbone}} & \multirow{2}{*}{\textbf{Res.}} & \multicolumn{2}{c|}{\textbf{Split \#1}} & \multicolumn{2}{c|}{\textbf{Split \#2}} & \multicolumn{2}{c|}{\textbf{Split \#3}} & \multicolumn{4}{c}{\textbf{Avg.}} \\
 & & & {Seen} & \textbf{Unseen} & {Seen} & \textbf{Unseen} & {Seen} & \textbf{Unseen} & {Seen} & \textbf{Unseen} & MACs & FPS \\ \hline
Balntas \textit{et al.}~\citep{balntas2017pose} & Wohlhart \textit{et al.}~\citep{wohlhart2015learning} & $64^2$ & 19.2 & 9.3 & 23.1 & 5.1 & 15.0 & 5.1 & 19.1 & 6.5 & 11.9M & 493\\
Wohlhart \textit{et al.}~\citep{wohlhart2015learning} & Wohlhart \textit{et al.}~\citep{wohlhart2015learning} & $64^2$ & 18.3 & 8.2 & 21.9 & 7.5 & 17.6 & 7.6 & 19.5 & 7.8 & 11.9M & 493\\
MPL~\citep{sundermeyer2020multi} & MPL~\citep{sundermeyer2020multi} & $128^2$ & 31.3 & 18.6 & 34.5 & 15.9 & 29.2 & 17.7 & 31.7 & 17.4 & 2.1B & 52.4\\
Zhao \textit{et al.}~\citep{zhao2022fusing} & ResNet~\citep{he2016deep} & $128^2$ & 54.9 & 40.1 & 63.4 & 32.7 & 49.9 & 37.5 & 56.1 & 36.8 & 1.7B & 43.8\\
Template-pose~\citep{nguyen2022templates} & ResNet~\citep{he2016deep} & $512^2$ & 78.5 & 73.6 & 85.2 & 73.1 & 78.0 & 87.2 & 80.6 & 77.9 & 130B & 46.5 \\
Template-pose~\citep{nguyen2022templates} & ResNet~\citep{he2016deep} & $224^2$ & 77.3 & 71.4 & 84.1 & 72.7 & 76.8 & 85.3 & 79.4 & 76.3 & 24.9B & 136\\\hline

Ours (VA) & SD-V1-5~\citep{rombach2022high} & $512^2$ & 80.2 & 79.4 & 87.2 & 78.0 & 80.2 & 95.6 & 82.5 & 84.6 & 902B & 10.2\\
Ours (NA) & SD-V1-5~\citep{rombach2022high} & $512^2$ & \textbf{81.9} & 81.4 & \textbf{88.5} & 79.0 & 79.5 & 96.0 & 83.3 & 85.5 & 929B & 9.5\\
Ours (CWA) & SD-V1-5~\citep{rombach2022high} & $512^2$ & 81.5 & {81.6} & 86.0 & 79.8 & 80.8 & \textbf{96.3} & 82.8 & \textbf{85.9} & 929B & 9.2\\ 
Ours (CAA) & SD-V1-5~\citep{rombach2022high} & $512^2$ & 81.7 & 80.8 & 86.8 & \textbf{80.0} & \textbf{81.9} & 95.7 & \textbf{83.5} & 85.5 & 719B & 14.9 \\
Ours (CAA) & SD-V1-5~\citep{rombach2022high} & $256^2$ & 81.6 & \textbf{81.8} & 86.2 & 78.3  & 80.8 & 94.8 & 82.9 & 85.0 & 179B & 42.2 \\
Ours (CAA) & SD-V1-5~\citep{rombach2022high} & $128^2$ & 81.4 & 81.3 & 86.0 & 78.3 & 79.9 & 94.6 & 82.4 & 84.7 & 44.9B & 62.6 \\
\hline
\end{tabular}}

\label{tab:lmo}
\end{table*}

We evaluate our method in Tables \ref{tab:lm} and \ref{tab:lmo}, and it achieves significantly better results compared to previous state-of-the-art methods~\citep{wohlhart2015learning,balntas2017pose,nguyen2022templates,sundermeyer2020multi,zhao2022fusing}.
In particular, on the unseen O-LM test set, our model achieves an average accuracy of 85.5\%, outperforming the template-pose method~\citep{nguyen2022templates} by 9.2\%. This demonstrates the superior generalizability of our proposed solution.
Wohlhart \textit{et al}.~\citep{wohlhart2015learning}, Balntas \textit{et al}.~\citep{balntas2017pose}, and MLP~\citep{sundermeyer2020multi} use carefully designed networks to learn feature embeddings with pose and class discriminative capabilities through the design of triplet and pair loss functions. These methods use global representations and are more likely to fail when objects are occluded and have limited generalizability to unseen objects.
Zhao \textit{et al.}~\citep{zhao2022fusing} use feature map representations and develop a similarity fusion module to predict the similarity between two features. Although this method improves the generalizability to unseen objects, its performance is not promising in occluded situations.
Template-pose~\citep{nguyen2022templates}, which uses InfoNCE~\citep{oord2018representation} loss and local per-patch representations masked by template masks, achieves better performance in the presence of occlusion. However, there is a significant performance gap between seen and unseen objects, with 99.1\% accuracy on Seen LM and 93.5\% on Unseen LM.
In contrast, our method significantly reduces this gap, achieving 99.6\% on Seen LM and 97.7\% on Unseen LM. 
This indicates that our method generalizes well to unseen objects, showcasing the efficacy of our proposed solution.

Tables \ref{tab:lm} and \ref{tab:lmo} present a comparative evaluation of the proposed aggregation methods on the LM and O-LM datasets. The results demonstrate that the nonlinear aggregation network (NA) consistently outperforms the vanilla aggregation network (VA) across nearly all datasets. This improvement can be attributed to the enhanced flexibility of NA, which incorporates nonlinearity to better adapt to the specific requirements of downstream tasks, enabling improved performance. Furthermore, significant performance gains are observed on datasets with unseen objects when the context-aware weight aggregation (CWA) method is employed. Unlike VA and NA, CWA dynamically adjusts aggregation weights based on the input context, allowing it to generalize more effectively to unseen data. This ability to adapt to varying input characteristics underlines the superior generalization capabilities of CWA, making it particularly effective in handling diverse and challenging scenarios.

Additionally, the confidence adaptive aggregation (CAA) method demonstrates competitive performance with notable computational efficiency. As shown in Table \ref{tab:lm}, CAA achieves an average accuracy of 99.6\% on seen objects and 97.7\% on unseen objects on the LM dataset, which is comparable to CWA. Crucially, CAA requires fewer MACs, 678B compared with 902B for VA and 929B for CWA, while achieving a higher throughput of 15.5 frames per second on a single A6000 GPU. This substantial reduction in computational cost, together with the improved inference throughput, makes CAA well suited for near real-time applications.


For the O-LM dataset (Table \ref{tab:lmo}), CAA achieves an average accuracy of 83.5\% on seen objects and 85.5\% on unseen objects, comparable to CWA. However, CAA exhibits a longer average inference time on the O-LM dataset than on the LM dataset. This is attributed to the higher prevalence of occluded scenes in O-LM, which necessitate processing more features by the aggregation network. This observation underscores ability of CAA to effectively allocate computational resources based on input complexity, further demonstrating its adaptability and efficiency. 

\todoo{Since practical deployments often operate under varying computational constraints, we further evaluate CAA at reduced input resolutions ($256$ and $128$). While lowering the resolution naturally decreases absolute accuracy to some extent, CAA maintains strong performance and achieves substantial improvements in efficiency, reaching 45.3 FPS and 68.1 FPS, respectively. In contrast, template-pose yields only marginal accuracy gains at higher resolutions despite requiring significantly more computation, resulting in a less favorable accuracy-efficiency trade-off. These observations confirm that CAA preserves real-time performance across a wide range of resource settings while maintaining its advantage over prior approaches.
}

\begin{figure}[!t]
\centering
\footnotesize
\includegraphics[width=0.45\textwidth]{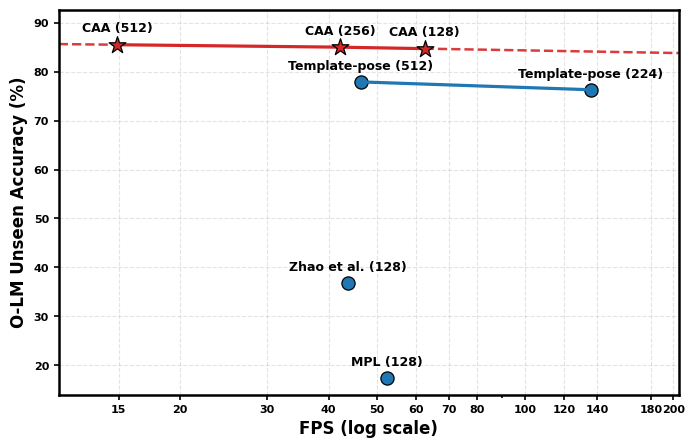}
\caption{\todoo{Comparison of accuracy-efficiency trade-offs across methods on the \emph{unseen} O-LM, illustrating the relationship between inference speed (FPS) and pose estimation accuracy. Numbers in parentheses indicate the input resolution used for each method.}}
\label{fig:tradeoff}
\end{figure}

\todoo{To further characterize the accuracy-efficiency trade-off of our approach, we visualize the relationship between inference speed and O-LM unseen accuracy in Figure~\ref{fig:tradeoff}. Prior methods generally fail to provide a satisfactory balance between accuracy and computational cost, with MPL~\citep{sundermeyer2020multi} and Zhao \textit{et al.}~\citep{zhao2022fusing} performing poorly on both axes. Template-Pose~\citep{nguyen2022templates} provides a better trade-off than earlier methods, but its balance between accuracy and efficiency still falls short. In contrast, CAA consistently forms the Pareto-optimal frontier across all resolutions. The 128 and 256 input-resolution variants retain accuracy comparable to the 512 model while delivering substantially higher FPS.}

These results confirm that the proposed aggregation methods, including VA, NA, CWA, and particularly CAA, substantially enhance the performance of aggregation networks. Among these, CAA achieves an optimal balance between high accuracy, computational efficiency, and adaptability, making it a practical choice for resource-constrained and real-time applications.

\subsubsection{T-LESS}\label{sub:com-tless}

The T-LESS dataset consists of particularly challenging texture-less rigid objects in highly cluttered scenes. In Table \ref{tab:tless}, we present the results of our method compared to state-of-the-art approaches~\citep{sundermeyer2018implicit,sundermeyer2020multi,nguyen2022templates}. Our method outperforms all the methods in this comparison. Notably, our method achieves higher accuracy with fewer templates compared to template-pose~\citep{nguyen2022templates}, 70.82\% vs. 58.87\%. When template-pose~\citep{nguyen2022templates} uses the same number of templates as our method, we outperform it by 11.53\% for seen objects and 14.05\% for unseen objects. This demonstrates the enhanced discriminative power of the features obtained from the diffusion model. Notably, our method can even achieve superior results for unseen objects compared to seen objects, highlighting the strong generalizability of our method. Additionally, our Confidence Adaptive Aggregation (CAA) method demonstrates significant computational efficiency. Despite a minor accuracy reduction of only 0.21\%, it reduces computational consumption by 26\%.

\todoo{To provide a more comprehensive comparison, we also evaluate a higher resolution ($512$) version of template-pose, but its performance remains significantly below ours (59.59\% vs. 70.82\%). Moreover, the $256$ and $128$ variants of CAA achieve competitive accuracy (70.31\% and 69.52\%) with dramatically lower MACs, demonstrating that our approach maintains strong performance even under substantially reduced computational budgets, further underscoring its practical efficiency for real-world deployment.}

\begin{table}[htp]
\centering
\footnotesize
\setlength\tabcolsep{2.5pt}
\caption{Results on T-LESS~\citep{hodan2017t}. }

\resizebox{0.48\textwidth}{!}{
\begin{tabular}{lcccccc}
\hline
\multirow{2}{*}{\textbf{Method}} & \multicolumn{1}{l}{\multirow{2}{*}{\textbf{\begin{tabular}[c]{@{}l@{}}Number\\ templates\end{tabular}}}} & \multirow{2}{*}{\textbf{Res.}} & \multicolumn{3}{c}{\textbf{Recall VSD}} & \multirow{2}{*}{\textbf{MACs}} \\ \cline{4-6}
 & \multicolumn{1}{l}{} &  & \textbf{Obj. 1-18} & \textbf{Obj. 19-30} & \textbf{Avg.} &  \\ \hline
Implicit~\citep{sundermeyer2018implicit} &  $92 \mathrm{~K}$ & $128^2$ &  35.60 & 42.45 & 38.34 & 2.1B \\
MPL~\citep{sundermeyer2020multi} &  $92 \mathrm{~K}$ & $128^2$ &  35.25 & 33.17 & 34.42 & 2.1B \\
Template-pose~\citep{nguyen2022templates} & $92 \mathrm{~K}$ & $224^2$ &  59.62 & 57.75 & 58.87 & 24.9B \\
\hline 
Template-pose~\citep{nguyen2022templates}  & $21 \mathrm{~K}$ & $512^2$ & 61.11 & 57.97 & 59.59 & 130B \\
Template-pose~\citep{nguyen2022templates}  & $21 \mathrm{~K}$ & $224^2$ &  59.14 & 56.91 & 58.25 & 24.9B \\
Ours (CWA) & $21 \mathrm{~K}$ & $512^2$ & \textbf{70.87} & \textbf{71.26} & \textbf{71.03} & 929B \\ 
Ours (CAA) & $21 \mathrm{~K}$ & $512^2$ & 70.67 & 70.96 & 70.82 & 682B \\
Ours (CAA) & $21 \mathrm{~K}$ & $256^2$ & 70.21 & 70.35 & 70.31 & 167B \\ 
Ours (CAA) & $21 \mathrm{~K}$ & $128^2$ & 69.82 & 69.21 & 69.52 & 44.7B \\ \hline
\end{tabular}
}
\label{tab:tless}
\end{table}

\subsubsection{\todoo{BOP}}


\todoo{For BOP Challenge evaluation, we integrate the diffusion features together with the proposed CAA module into the GigaPose~\citep{nguyen2024gigapose} pipeline. Specifically, we use the original $224 \times 224$ input resolution and replace the DINO-based~\citep{oquab2023dinov2} features with diffusion features aggregated by our CAA module. To ensure compatibility, the extractor in CAA is adjusted so that the aggregated multi-scale diffusion features are downsampled through modified convolutional layers to a $16 \times 16$ representation with the same channel dimensionality as the original encoder. During training, both the diffusion U-Net and the VAE of the Stable Diffusion model are kept frozen, and only the CAA module is trained jointly with the rest of the GigaPose pipeline. The training objective follows the standard GigaPose loss, augmented with a confidence term $\mathcal{L}_{\text{conf}}$ to encourage pose-discriminative feature aggregation. For fairness, all compared methods use the same CNOS~\citep{nguyen2023cnos} detections and segmentation masks, and all refinement results are produced using the standard MegaPose~\citep{labbe2022megapose} refiner. This ensures that any performance differences arise solely from the feature representation and the proposed CAA aggregation. The only exception is GenFlow~\citep{moon2024genflow}, for which we report the official results because its implementation is not publicly available.
}


\todoo{Across the seven core BOP datasets, combining diffusion features with the proposed CAA module \todooo{improves the average AR over GigaPose}, as shown in Table~\ref{tab:bop}. Notably, it attains higher single-hypothesis and five-hypothesis AR than GigaPose while using substantially fewer trainable parameters (70.7M vs.\ 360M). The advantages are especially evident on challenging datasets such as ICBIN and YCB-V, where the proposed aggregation produces more reliable pose predictions under significant variability. These results demonstrate that strong performance can be achieved
through parameter-efficient task-specific adaptation while keeping the
pretrained diffusion backbone frozen.}

\todooo{
For completeness, the loaded Stable Diffusion model contains approximately 1.080B parameters. The full BOP pipeline reaches a maximum stage-wise peak GPU memory usage of 9.27 GiB. The onboarding cost is approximately 3.4 s per object, well below the BOP limit of 300 s per object. Under the five-hypothesis setting, the end-to-end inference time is approximately 7.2 s per image, comprising 0.21 s for detection and retrieval, 0.63 s for coarse pose estimation, and 6.36 s for refinement and scoring.
}


\begin{table*}[t]
\centering
\caption{
\todoo{Comparison with state-of-the-art methods on the seven core BOP datasets.
We report Average Recall (AR) under the evaluation settings defined in the BOP Challenge, 
including both single- and multi-hypothesis predictions.}
}
\vspace{2mm}

\resizebox{\textwidth}{!}{
\begin{tabular}{lccccccccccc}
\toprule
Method  & Pose refine. & Trainable  Params & LM-O & T-LESS & TUD-L & ICBIN & ITODD & HB & YCB-V & Avg. & Time \\
\midrule
\multicolumn{12}{c}{\textbf{Pose estimation (single hypothesis)}} \\
\midrule
MegaPose~\citep{labbe2022megapose} & MegaPose~\citep{labbe2022megapose}  & 64.9 M & 49.9 & 47.7 & \textbf{65.3} & 36.7 & 31.5 & 65.4 & 60.1 & 50.9 & 31.7 s \\
FoundPose~\citep{ornek2024foundpose} & MegaPose~\citep{labbe2022megapose}  & 43.1 M & 55.4 & 51.0 & 63.3 & 43.0 & 34.6 & 69.5 & \textbf{66.1} & \underline{54.7} & 4.4 s \\
GigaPose~\citep{nguyen2024gigapose} & MegaPose~\citep{labbe2022megapose}  & 360 M & \underline{55.6} & \textbf{54.6} & 57.8 & \underline{44.3} & \textbf{37.8} & \underline{69.6} & 63.4 & \underline{54.7} & 2.4 s \\
Ours & MegaPose~\citep{labbe2022megapose}  & 70.7 M & \textbf{56.5} & \underline{53.6} & \underline{62.2} & \textbf{45.1} & \underline{36.0} & \textbf{70.3} & \underline{64.6} & \textbf{55.5} & \textbf{2.2 s} \\

\midrule
\multicolumn{12}{c}{\textbf{Pose estimation (five hypotheses)}} \\
\midrule

MegaPose~\citep{labbe2022megapose} & MegaPose~\citep{labbe2022megapose} & 64.9 M & 56.0 & 50.7 & \textbf{68.4} & 41.4 & 33.8 & 70.4 & 62.1 & 54.7 & 47.4 s \\
FoundPose~\citep{ornek2024foundpose} & MegaPose~\citep{labbe2022megapose}  & 43.1 M & 58.6 & 54.9 & \underline{65.7} & 44.4 & 36.1 & 70.3 & \underline{67.3} & 56.8 & 11.2 s \\
GenFlow~\citep{moon2024genflow} & GenFlow~\citep{moon2024genflow} & - &  56.3 & 52.3 &  \textbf{68.4}  & 45.3 &  \underline{39.5} &  \textbf{73.9}  & 63.3 &  57.1 & 20.9 s \\
GigaPose~\citep{nguyen2024gigapose} & MegaPose~\citep{labbe2022megapose} & 360 M & \underline{59.9} & \textbf{57.0} & 64.5 & \underline{46.7} & \textbf{39.7} & \underline{72.2} & 66.3 & \underline{57.9} & 7.3 s \\
Ours & MegaPose~\citep{labbe2022megapose}  & 70.7 M & \textbf{60.4} & \underline{55.9} & 65.3 & \textbf{47.9} & 37.5 & {73.2} & \textbf{68.2} & \textbf{58.3} & \textbf{7.2 s} \\

\bottomrule
\end{tabular}
}
\label{tab:bop}
\end{table*}

\subsection{Ablation study}
\label{sub:ablation}

\subsubsection{Timestep}

\begin{figure}[htb]
\centering
\footnotesize
\includegraphics[width=0.435\textwidth]{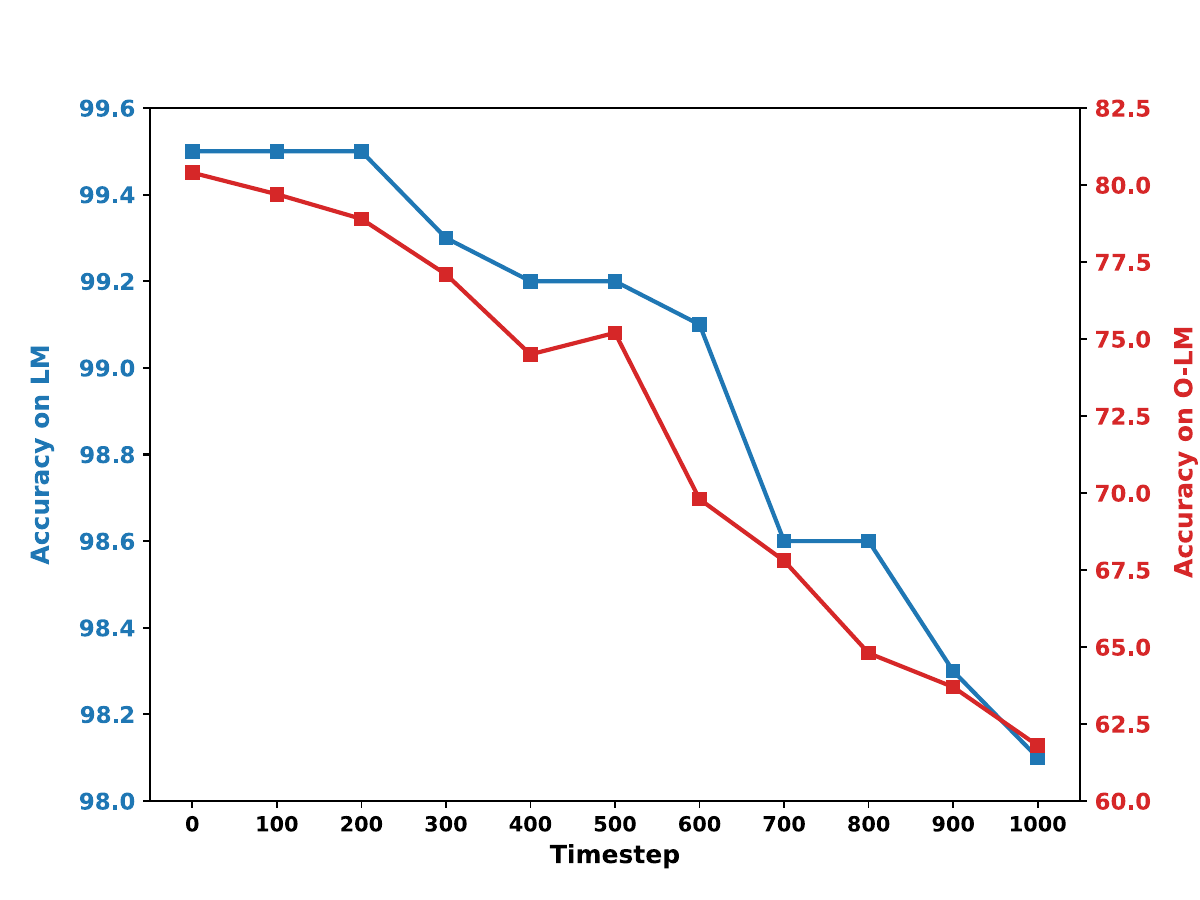}
\caption{Ablation on timestep $t$. The accuracy is measured for the model by extracting features from Stable Diffusion at different timesteps on LM and O-LM datasets.}
\label{fig:time_step}
\end{figure}

As mentioned in Sect.~\ref{subsec3-2}, the diffusion model encodes more specific information at various timesteps.
To understand the impact of timestep, 
we conduct a performance comparison by extracting features from the  SD model at different timesteps. 
We employ our vanilla aggregation network to conduct experiments on objects included in split \#1 of both Seen LM and Seen O-LM to quickly identify the best timestep. We sampled eleven timesteps from 0 to 1000, evenly spaced.
The results on seen objects of split 1 of the LM and O-LM datasets are presented in Fig. \ref{fig:time_step}. The results show that the accuracies on LM and O-LM decrease as the timestep increases. This is because the diffusion model assumes less noise in the input when the timestep is small, which matches the scenarios in our images.
Therefore, we set the timestep to 0 for the remaining experiments.

\subsubsection{Other models pre-trained at large scale}

In this section, we compare the features of other \emph{large} models. 
We first use another SD model, SDv2-0, which has the \emph{same network architecture} as SDv1-5 but is trained on \emph{different data}. Same as SDv1-5, we set the timestep to 0 and extract features from all layers in the UNet. Furthermore, our method can be easily incorporated into other pre-trained large visual models. As shown in Tables \ref{tab:lm-large} and \ref{tab:lmo-large}, we have also applied our context-aware weight aggregation network to other models, including DINOv2~\citep{oquab2023dinov2} pre-trained on LVD-142M dataset and OpenCLIP~\citep{Radford2021LearningTV} pre-trained on LAION~\citep{schuhmann2022laion} dataset. It is worth noting that OpenCLIP uses the \emph{same training data} as SDv1-5 but employs a \emph{different network architecture}.
We use the best checkpoints available on their GitHub repositories, ViT-G for DINOv2 and OpenCLIP. Similar to Stable Diffusion, we extract features from all transformer blocks of these models. We aggregate the intermediate features using the context-aware aggregation network. The results are shown in Tables \ref{tab:lm-large} and \ref{tab:lmo-large}. We can see that Stable Diffusion v1-5 achieves the best results in most of the benchmarks. This indicates that the intermediate features of the diffusion model have superior capabilities compared to the discriminative models for object pose estimation tasks.


\begin{table}[htp]
\centering
\setlength\tabcolsep{2.5pt}
\caption{{Comparison with other large models on LM~\citep{hinterstoisser2013model}. }}
\resizebox{\linewidth}{!}{\begin{tabular}{l|cc|cc|cc|cc}
\hline
\multirow{2}{*}{\textbf{Backbone}} & \multicolumn{2}{c|}{\textbf{Split \#1}} & \multicolumn{2}{c|}{\textbf{Split \#2}} & \multicolumn{2}{c|}{\textbf{Split \#3}} & \multicolumn{2}{c}{\textbf{Avg.}} \\
 & {Seen} & \textbf{Unseen} & {Seen} & \textbf{Unseen} & {Seen} & \textbf{Unseen} & {Seen} & \textbf{Unseen} \\ \hline
SD-V1-5~\citep{rombach2022high} & 99.5 & \textbf{98.3} & 99.6 & 98.9 & \textbf{99.8} & 96.3 & \textbf{99.7} & 97.9 \\
SD-V2-0~\citep{rombach2022high} & 99.6 & 98.1 & 99.4 & \textbf{99.5} & 99.6 & 97.5 & 99.5 & \textbf{98.4} \\
OpenCLIP~\citep{radford2021learning} & 99.4 & 97.0 & \textbf{99.7} & 98.5 & 99.7 & 95.9 & 99.6 & 97.1 \\
DINOv2~\citep{oquab2023dinov2} & \textbf{99.7} & 96.6 & 99.6 & 99.1 & 99.5 & \textbf{97.8} & 99.6 & 97.8 \\ \hline
\end{tabular}}

\label{tab:lm-large}
\end{table}

\begin{table}[htp]
\centering
\setlength\tabcolsep{2.5pt}
\caption{{Comparison with other large models on O-LM~\citep{brachmann2014learning}.}}
\resizebox{\linewidth}{!}{\begin{tabular}{l|cc|cc|cc|cc}
\hline
\multirow{2}{*}{\textbf{Backbone}} & \multicolumn{2}{c|}{\textbf{Split \#1}} & \multicolumn{2}{c|}{\textbf{Split \#2}} & \multicolumn{2}{c|}{\textbf{Split \#3}} & \multicolumn{2}{c}{\textbf{Avg.}} \\
 & {Seen} & \textbf{Unseen} & {Seen} & \textbf{Unseen} & {Seen} & \textbf{Unseen} & {Seen} & \textbf{Unseen} \\ \hline
SD-V1-5~\citep{rombach2022high} & 81.5 & \textbf{81.6} & 86.0 & \textbf{79.8} & \textbf{80.8} & \textbf{96.3} & 82.8 & \textbf{85.9} \\ 
SD-V2-0~\citep{rombach2022high}      & 82.9 & 81.0 & \textbf{88.1} & 79.1 & 79.2 & 95.5 & \textbf{83.4} & 85.2 \\
OpenCLIP~\citep{radford2021learning} & 79.8 & 76.5 & 85.7 & 73.9 & 80.1 & 91.9 & 81.9 & 81.1 \\
DINOv2~\citep{oquab2023dinov2}       & \textbf{84.1} & 73.7 & 96.1 & 74.3 & 79.0 & 86.3 & 83.1 & 78.1 \\ \hline
\end{tabular}}

\label{tab:lmo-large}
\end{table}

\subsubsection{Confidence threshold}\label{subsec:conf}


To evaluate the impact of the confidence threshold $\delta$ on both accuracy and speed, we compare results on the LM and O-LM datasets, varying the threshold values $\delta = \{0.65, 0.7, 0.75\}$. 
\todooo{We first use Split \#1 as the threshold-selection split and select $\delta=0.7$ according to the observed accuracy--efficiency trade-off. We then keep this threshold fixed and apply it to Splits \#2 and \#3 without further tuning. The same threshold also maintains a favorable accuracy--efficiency trade-off on these two splits, supporting the stability of the selected threshold across different object splits.}

As shown in Table \ref{tab:conf}, increasing the confidence threshold improves accuracy but also increases computational costs. Beyond a certain point, although computational costs continue to rise, accuracy plateaus. This is a reasonable observation, as certain images require the aggregation of more features to achieve optimal performance. However, once this limit is reached, adding more features does not further enhance performance. 
\todooo{In particular, increasing $\delta$ from $0.65$ to $0.7$ provides a clear improvement in accuracy, whereas further increasing it from $0.7$ to $0.75$ yields only marginal gains while noticeably increasing computational cost.}
Therefore, \todooo{based on the threshold selection on Split \#1 and its consistent performance on Splits \#2 and \#3,} we set $\delta = 0.7$ in our experiments.

\begin{table}[htbp]
\centering
\caption{Impact of confidence threshold $\delta$.}
\begin{tabular}{c|cc|cc|cc}
\hline
\multirow{2}{*}{$\delta$} & \multicolumn{2}{c|}{Avg. on LM} & \multicolumn{2}{c|}{Avg. on O-LM} & \multicolumn{2}{c}{Avg.} \\
 & Seen & Unseen & Seen & Unseen & MACs & FPS \\ \hline
0.65 & 99.4 & 97.1 & 82.5 & 83.0 & 610B & 17.6 \\
0.7 & 99.6 & 97.7 & 83.5 & 85.5 & 719B & 14.9 \\
0.75 & 99.7 & 97.9 & 83.6 & 85.6 & 863B & 12.4 \\ \hline
\end{tabular}

\label{tab:conf}
\end{table}

\begin{figure}[htb]
\centering
\footnotesize
\subfloat{
\begin{minipage}[t]{0.190\linewidth}
    \centering
    \includegraphics[width=0.62in]{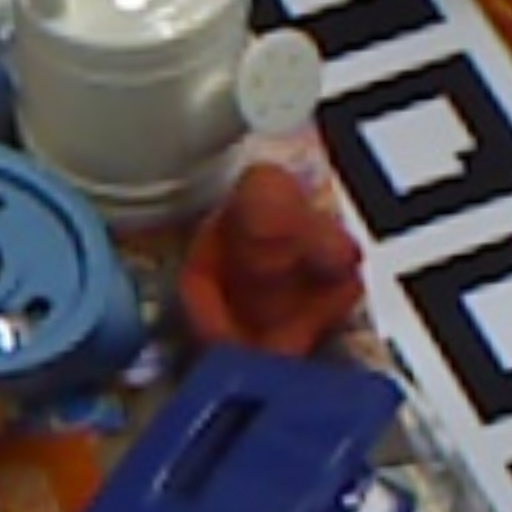}\\
    \vspace{0.004cm}
    \includegraphics[width=0.62in]{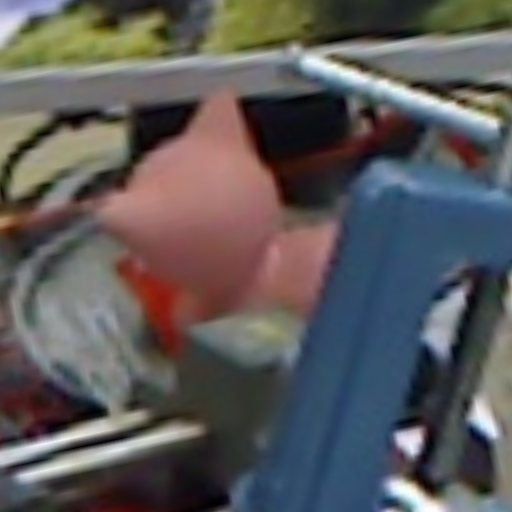}\\
    \vspace{0.004cm}
    \includegraphics[width=0.62in]{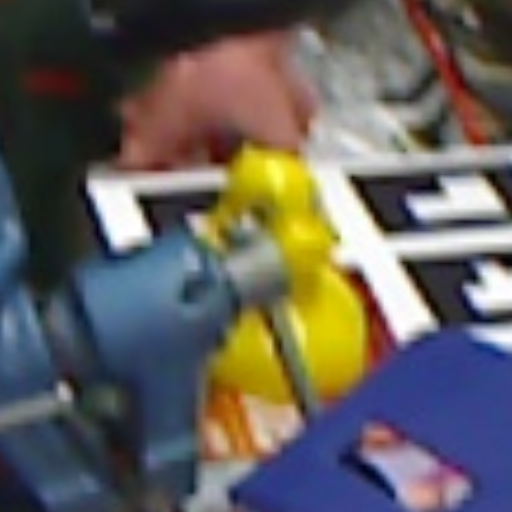}\\
    \vspace{0.004cm}
    \includegraphics[width=0.62in]{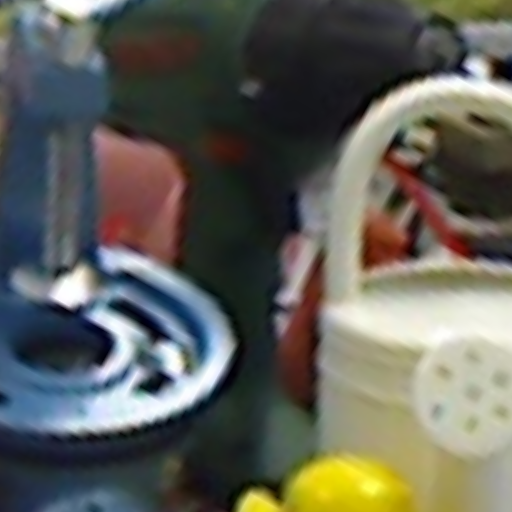}\\
    \vspace{0.004cm}
    \includegraphics[width=0.62in]{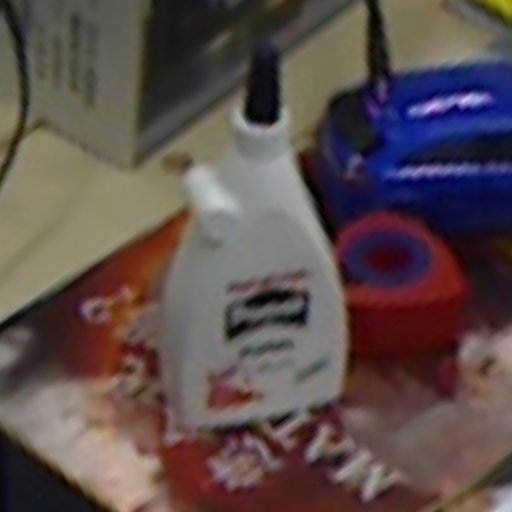}\\
    \vspace{0.004cm}
    \includegraphics[width=0.62in]{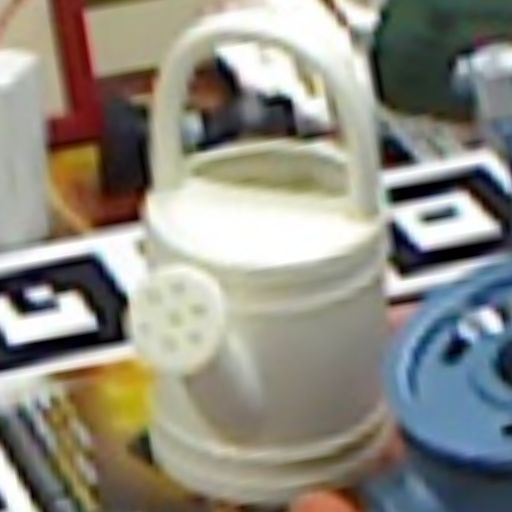}\\
    \centerline{Query}
\end{minipage}%
\begin{minipage}[t]{0.190\linewidth}
    \centering
    \includegraphics[width=0.62in]{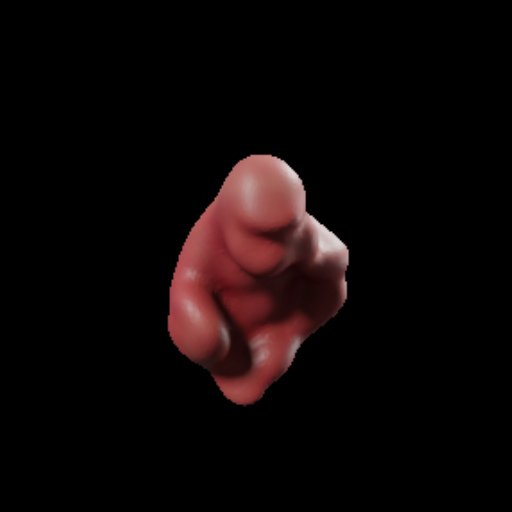}\\
    \vspace{0.004cm}
    \includegraphics[width=0.62in]{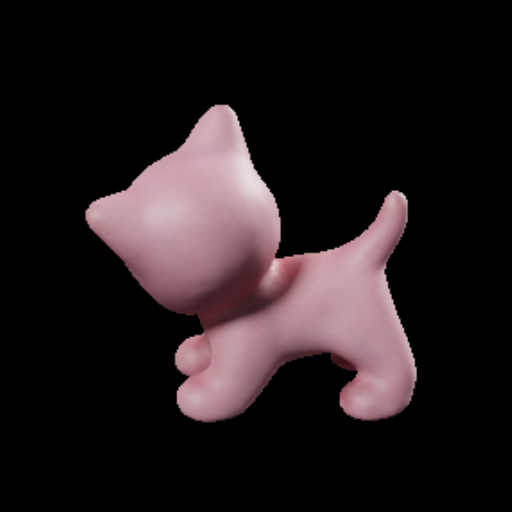}\\
    \vspace{0.004cm}
    \includegraphics[width=0.62in]{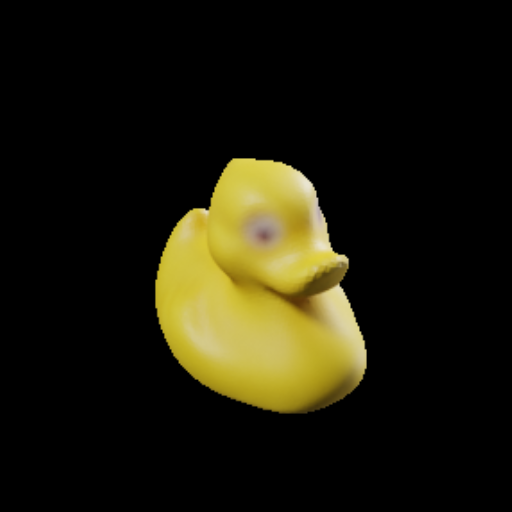}\\
    \vspace{0.004cm}
    \includegraphics[width=0.62in]{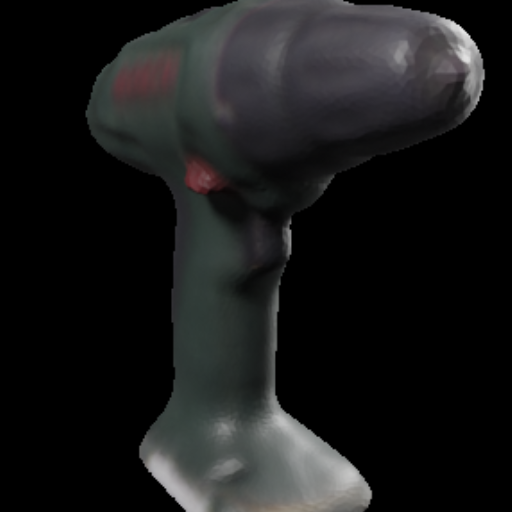}\\
    \vspace{0.004cm}
    \includegraphics[width=0.62in]{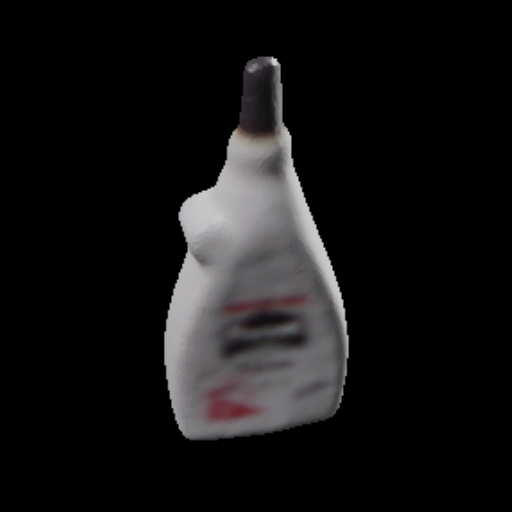}\\
    \vspace{0.004cm}
    \includegraphics[width=0.62in]{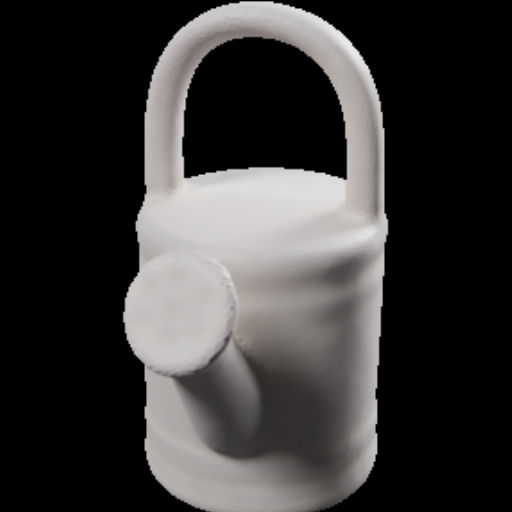}\\
    \centerline{GT}
\end{minipage}%
\begin{minipage}[t]{0.190\linewidth}
    \centering
    \includegraphics[width=0.62in]{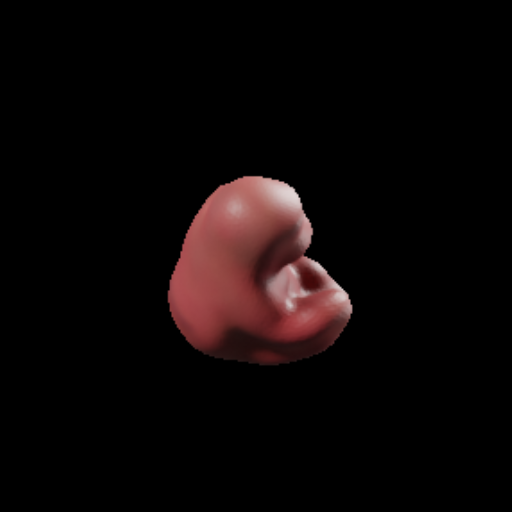}\\
    \vspace{0.004cm}
    \includegraphics[width=0.62in]{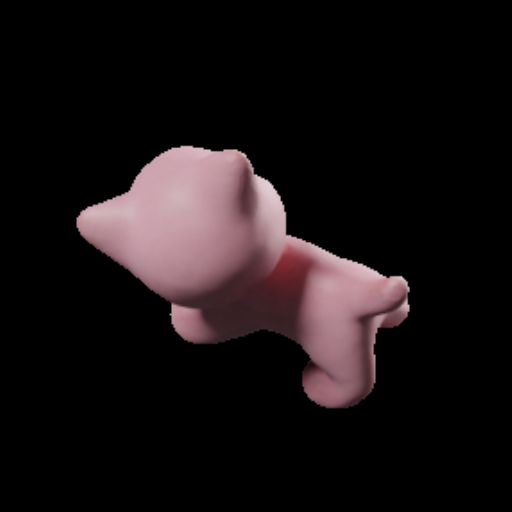}\\
    \vspace{0.004cm}
    \includegraphics[width=0.62in]{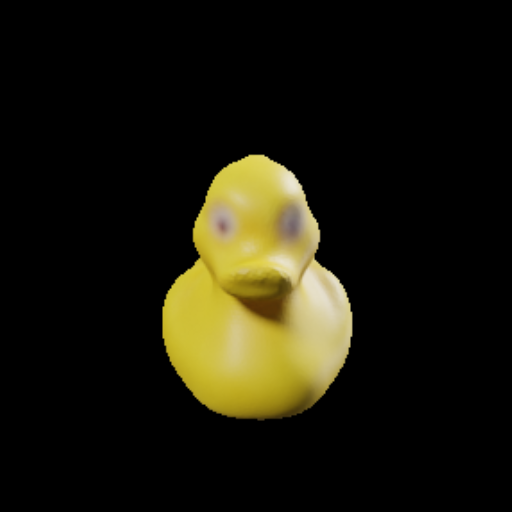}\\
    \vspace{0.004cm}
    \includegraphics[width=0.62in]{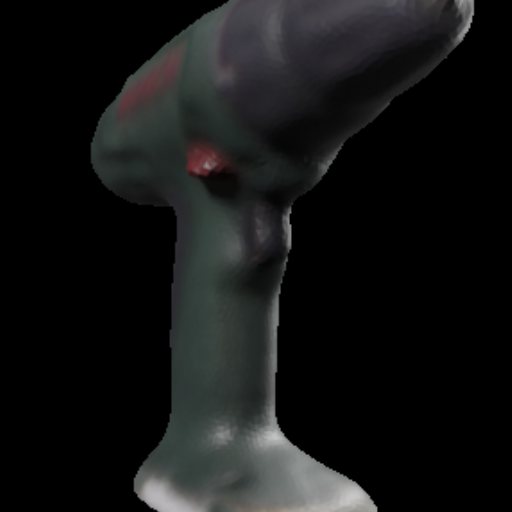}\\
    \vspace{0.004cm}
    \includegraphics[width=0.62in]{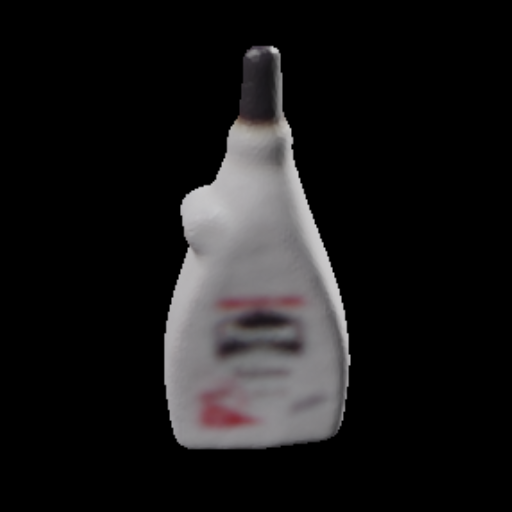}\\
    \vspace{0.004cm}
    \includegraphics[width=0.62in]{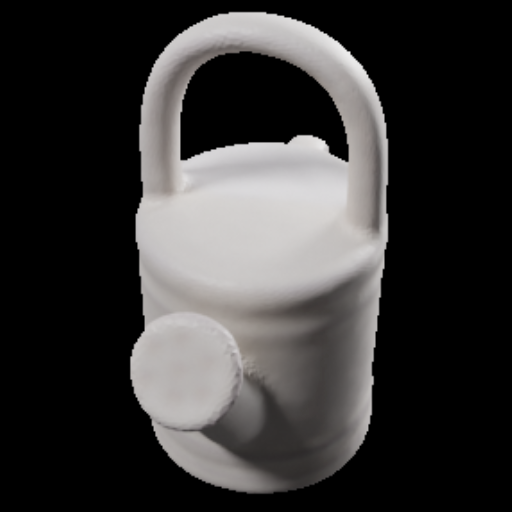}\\
    \centerline{$\delta=0.65$}
\end{minipage}%
\begin{minipage}[t]{0.190\linewidth}
    \centering
    \includegraphics[width=0.62in]{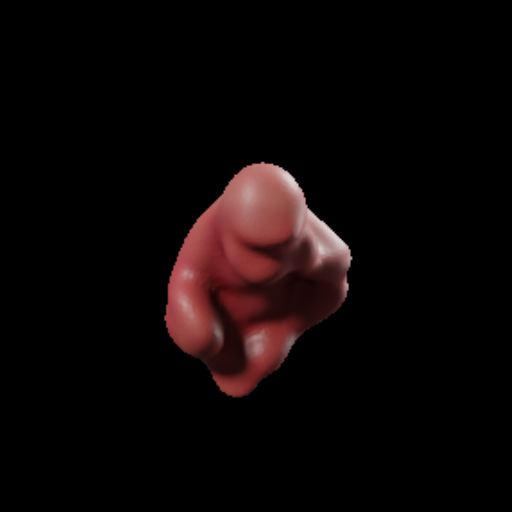}\\
    \vspace{0.004cm}
    \includegraphics[width=0.62in]{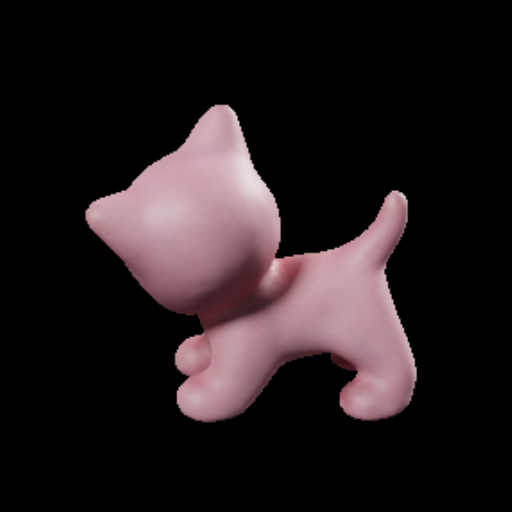}\\
    \vspace{0.004cm}
    \includegraphics[width=0.62in]{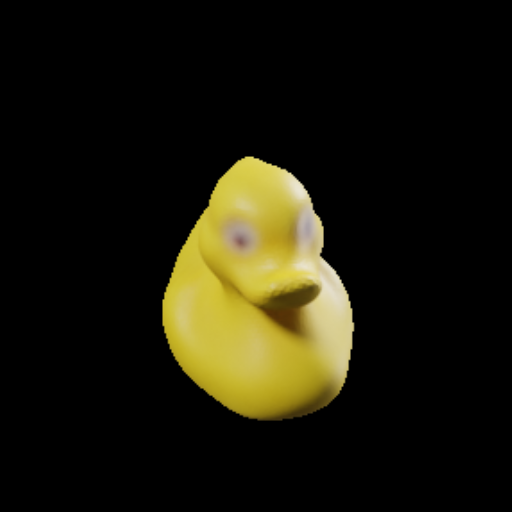}\\
    \vspace{0.004cm}
    \includegraphics[width=0.62in]{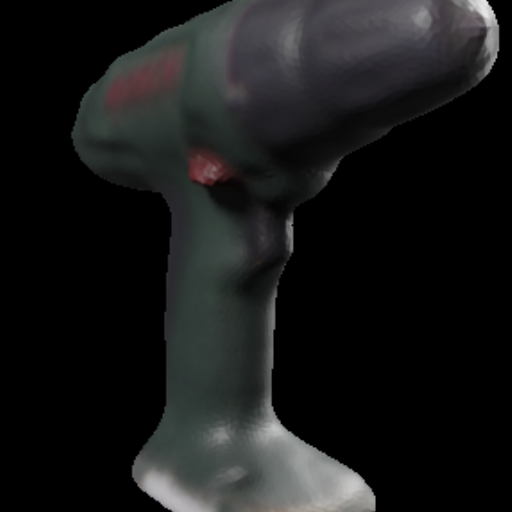}\\
    \vspace{0.004cm}
    \includegraphics[width=0.62in]{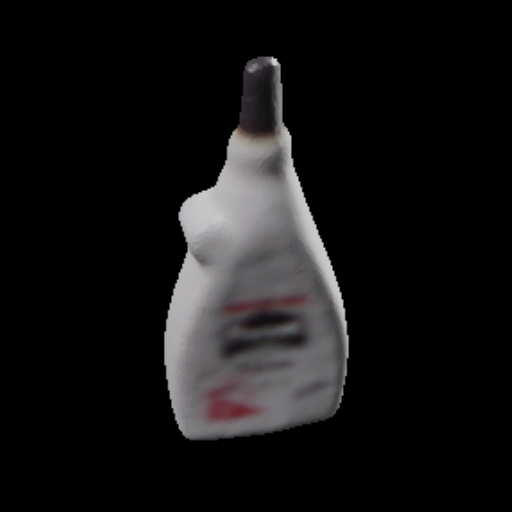}\\
    \vspace{0.004cm}
    \includegraphics[width=0.62in]{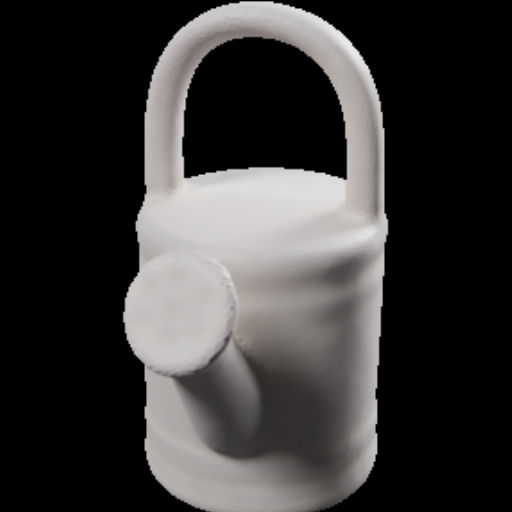}\\
    \centerline{$\delta=0.7$}
\end{minipage}%
\begin{minipage}[t]{0.190\linewidth}
    \centering
    \includegraphics[width=0.62in]{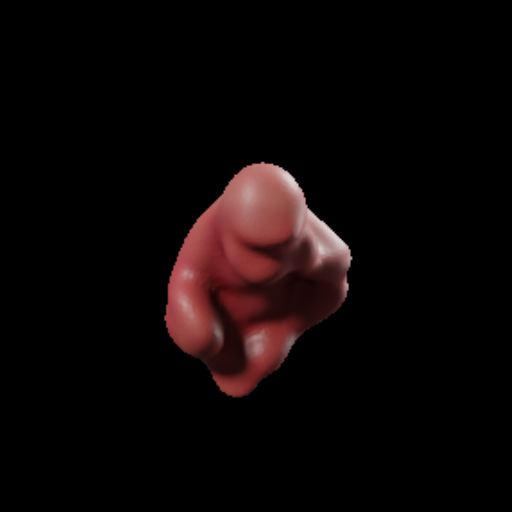}\\
    \vspace{0.004cm}
    \includegraphics[width=0.62in]{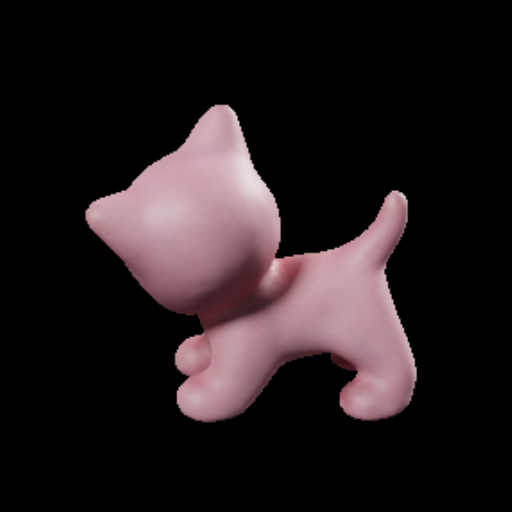}\\
    \vspace{0.004cm}
    \includegraphics[width=0.62in]{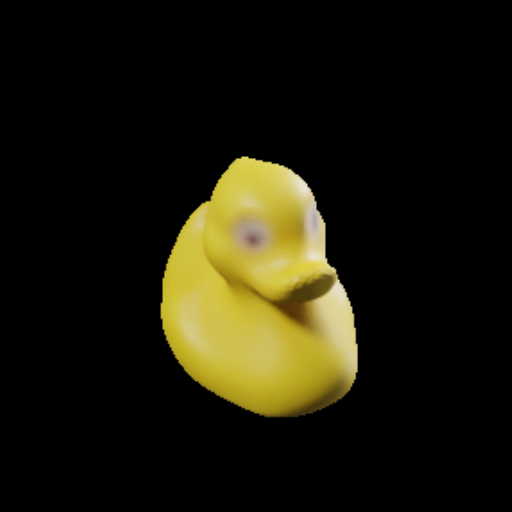}\\
    \vspace{0.004cm}
    \includegraphics[width=0.62in]{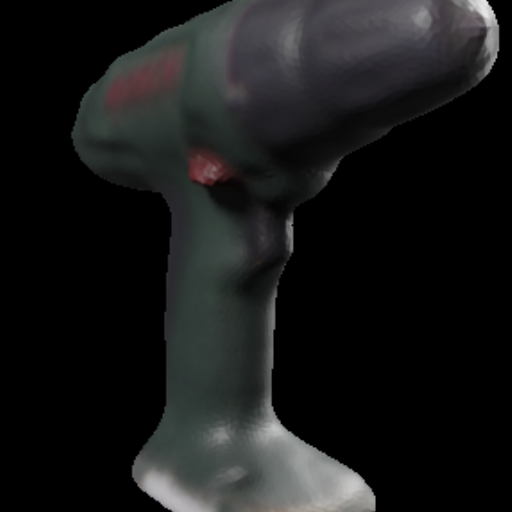}\\
    \vspace{0.004cm}
    \includegraphics[width=0.62in]{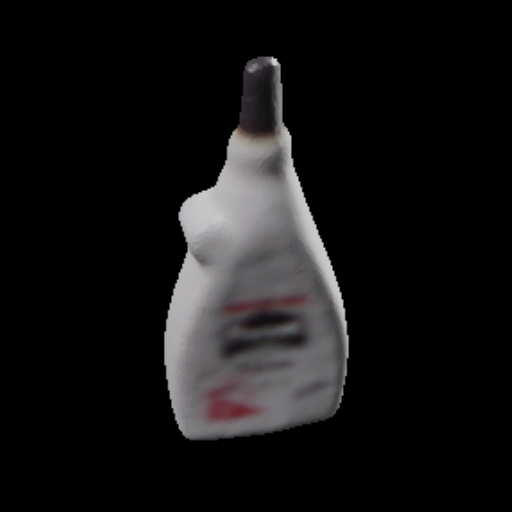}\\
    \vspace{0.004cm}
    \includegraphics[width=0.62in]{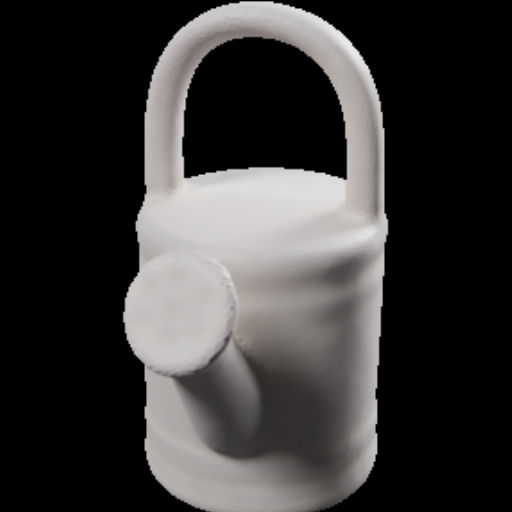}\\
    \centerline{$\delta=0.75$}
\end{minipage}%
}
\centering
\caption{Visualization of the impact of varying confidence thresholds $\delta$.} 
\label{fig:vis_conf}
\end{figure}

To further illustrate the impact of varying confidence thresholds, we visualize the results on the unseen O-LM dataset for different values of $\delta$ in Fig. \ref{fig:vis_conf}. The visualizations clearly show how adjustments to the threshold influence accuracy. As the threshold increases, we observe a continuous improvement in accuracy, consistent with the data presented in Table \ref{tab:conf}. However, the visualizations also highlight a diminishing return in accuracy gains as the threshold reaches higher values. This trend supports our earlier observation that beyond a certain point, adding more features introduces redundancy and noise, which do not contribute to further accuracy improvements. These visualizations reinforce our decision to set $\delta = 0.7$ for our experiments, as this value achieves an optimal trade-off between maximizing accuracy and maintaining computational efficiency.

\subsubsection{\todooo{Confidence Reliability Analysis}}
The confidence score in CAA is trained using the feature similarity to the
corresponding positive template rather than binary retrieval correctness.
Nevertheless, during inference, it serves as the signal for determining
whether the current aggregated representation is sufficiently reliable for
early exiting. We therefore further evaluate whether the learned confidence
score empirically reflects actual retrieval reliability from three
complementary perspectives: discrimination of retrieval correctness,
empirical calibration, and the resulting exit behavior.

\paragraph{Discrimination of retrieval correctness.}
We first evaluate whether the confidence score is informative of actual
retrieval correctness. At each evaluated exit block, each query sample is
associated with (i) its predicted confidence score and (ii) a binary label
indicating whether the resulting retrieval is correct. We then use the
confidence score as the decision variable and compute the Area Under the
Receiver Operating Characteristic Curve (AUROC) by sweeping a decision
threshold over the confidence scores.

The ROC curve measures the trade-off between the true-positive rate
(correct retrievals identified as confident) and the false-positive rate
(incorrect retrievals identified as confident) at different confidence
thresholds. Accordingly, AUROC summarizes how well the confidence score
ranks correct retrievals above incorrect ones without relying on a particular
stopping threshold. An AUROC of $0.5$ corresponds to random discrimination,
whereas an AUROC of $1.0$ indicates perfect separation. Equivalently, a higher
AUROC means that a randomly selected correct retrieval is more likely to
receive a higher confidence score than a randomly selected incorrect retrieval.

Table~\ref{tab:caa_auroc} reports the AUROC at different aggregation depths on
LM. The confidence score consistently achieves AUROC values above $0.5$ across
all evaluated exits. For seen objects, the AUROC ranges from $0.736$ to $0.759$,
while for unseen objects it ranges from $0.683$ to $0.711$. The relatively
stable results across aggregation depths indicate that the learned confidence
provides a consistent signal for distinguishing retrieval outcomes throughout
the progressive aggregation process.

\begin{table}[htbp]
\centering
\small
\caption{AUROC of the CAA confidence score for discriminating correct and
incorrect retrievals at different exit blocks on LM. Higher is better.}
\label{tab:caa_auroc}
\begin{tabular}{c|cc}
\hline
Exit Block & LM Seen & LM Unseen \\
\hline
2 & 0.743 & 0.694 \\
4 & 0.752 & 0.701 \\
6 & 0.736 & 0.683 \\
8 & 0.759 & 0.711 \\
\hline
\end{tabular}
\end{table}

\paragraph{Calibration analysis.}
We further evaluate the empirical agreement between the confidence score and
actual retrieval correctness. Since Split \#1 is used to select the stopping
threshold $\delta=0.7$, we report the calibration results on Splits \#2 and
\#3, which are not involved in threshold selection. This avoids evaluating
calibration on the same split used for choosing the stopping criterion. We use
the confidence score predicted at the final aggregation block so that all
samples are evaluated at the same depth, avoiding the sample-selection effect
introduced by early exiting.

We use two complementary calibration metrics. Expected Calibration Error (ECE)
measures the average discrepancy between the predicted confidence and the
empirical retrieval accuracy over confidence bins. The Brier Score measures
the mean squared discrepancy between the confidence score and the binary
retrieval-correctness label. Lower values indicate better empirical agreement
with retrieval correctness.

As shown in Table~\ref{tab:caa_calibration}, CAA obtains an ECE of $0.006$ and
a Brier Score of $0.0101$ on seen objects, and an ECE of $0.026$ and a Brier
Score of $0.0229$ on unseen objects. These results indicate a low empirical
calibration error under our evaluation protocol.

\begin{table}[htbp]
\centering
\small
\caption{Empirical calibration of the CAA confidence score on LM using the
final-block confidence of Splits \#2 and \#3. Lower is better.}
\label{tab:caa_calibration}
\begin{tabular}{lcc}
\hline
Subset & ECE $\downarrow$ & Brier Score $\downarrow$ \\
\hline
LM Seen   & 0.006 & 0.0101 \\
LM Unseen & 0.026 & 0.0229 \\
\hline
\end{tabular}
\end{table}

\begin{figure}[htbp]
    \centering
    \includegraphics[width=0.45 \textwidth]{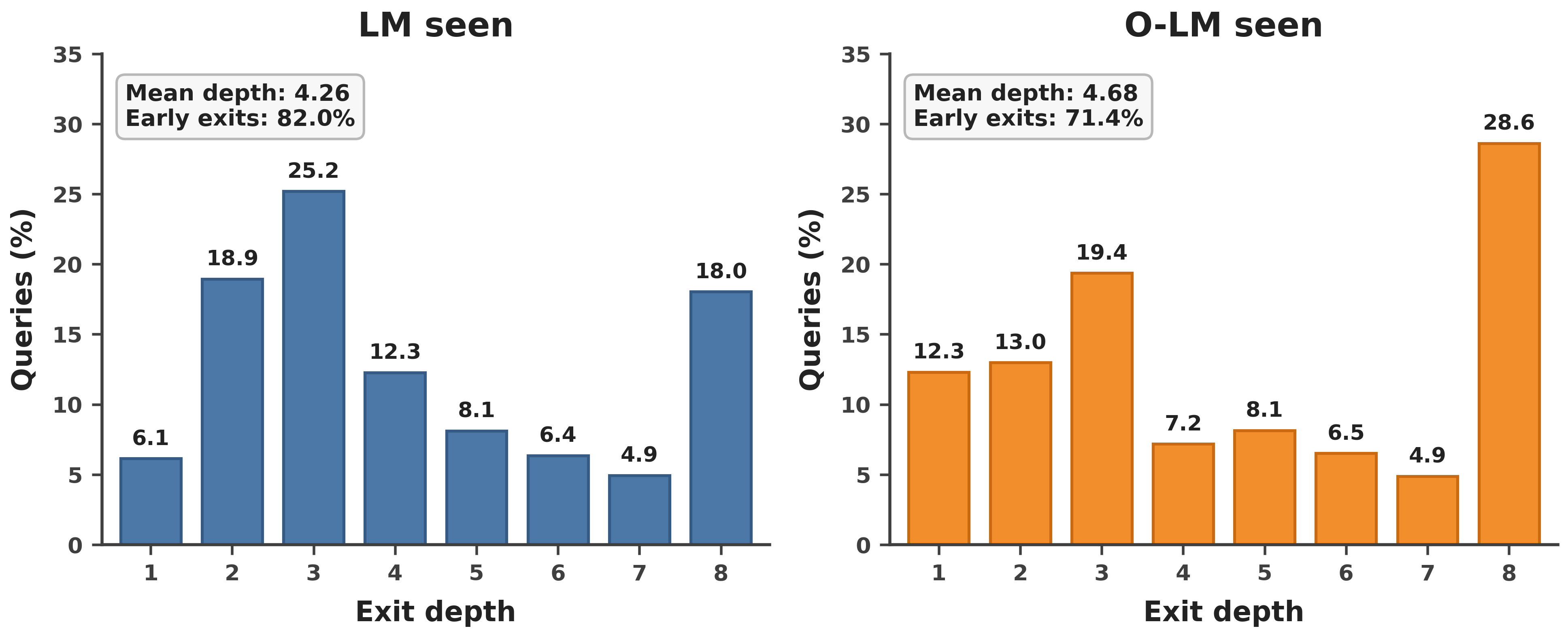}
    \caption{Exit-depth distributions of CAA under $\delta=0.7$ on LM and
    O-LM seen objects.}
    \label{fig:exit_depth_distribution}
\end{figure}

\paragraph{Exit-depth distribution.}
Finally, we analyze how the learned confidence score affects the actual
early-exit behavior under the selected threshold $\delta=0.7$.
Figure~\ref{fig:exit_depth_distribution} shows the exit-depth distributions on
LM and O-LM seen objects. On LM, the average exit depth is $4.26$, and $82.0\%$
of the samples exit before the final block. On O-LM, the average exit depth is
$4.68$, with $71.4\%$ of the samples exiting before the final block. These
results show that the confidence score produces sample-dependent exit
decisions rather than simply driving most samples to a fixed aggregation
depth.

Overall, the AUROC results show that the confidence score consistently
discriminates between correct and incorrect retrievals, while the ECE and
Brier Score provide complementary evidence of low empirical calibration error.
The exit-depth distributions further demonstrate that this confidence signal
is effectively used to produce adaptive, sample-dependent early exits.
Together, these analyses validate the learned confidence as an effective
retrieval-quality and stopping signal for CAA, while it remains a
similarity-supervised score rather than a directly supervised probability of
correct retrieval.

\subsubsection{Adaptive vs fixed exiting}\label{subsec:ada-vs-fix}


To highlight the advantages of confidence adaptive aggregation over fixed-budget aggregation, we conduct ablation experiments in two settings. In the first setting, we use the confidence adaptive aggregation network. In the second setting, we use a fixed-budget aggregation network, processing features from a predetermined number of blocks for each test. The results, obtained from the \emph{unseen} objects in the O-LM dataset, reveal that our confidence adaptive aggregation (CAA) method consistently outperforms the fixed-budget approach in terms of average accuracy. \todooo{Specifically, CAA consistently maintains higher accuracy than the corresponding fixed-budget setting, with improvements of $1.2$, $0.5$, and $0.4$ percentage points at average aggregation budgets of 2, 4, and 6 features, respectively, and limited variation across the three object splits.} Table \ref{tab:ada_vs_fix} illustrates these findings.

\begin{table}[htbp]
\centering
\footnotesize
\caption{\todooo{Impact of confidence adaptive exiting. Results are averaged over
three object splits. For each column, CAA* operates with the corresponding
average number of aggregated features. The $\pm$ values denote the sample
standard deviation of the paired CAA--Fixed (pp) accuracy differences across the
three splits.}}
\resizebox{0.46\textwidth}{!}{
\begin{tabular}{l|ccc}
\hline
No. of agg. features & 2 & 4 & 6 \\ \hline
Fixed budget & 82.0 & 85.0 & 85.3 \\
CAA* & 83.2 & 85.5 & 85.7 \\
CAA* $-$ Fixed (pp) & $+1.2 \pm 0.3$ & $+0.5 \pm 0.1$ & $+0.4 \pm 0.1$ \\
\hline
\end{tabular}
}

\label{tab:ada_vs_fix}
\end{table}

\subsubsection{\todooo{Controlled ablation of diffusion features and CAA}}
\label{subsec:bop_ablation}

To separate the effects of diffusion features and CAA, we conduct a controlled ablation study under the same BOP setting. The DINOv2 backbone is fine-tuned following GigaPose, whereas the Stable Diffusion backbone is kept frozen in all diffusion-feature configurations.

As shown in Table~\ref{tab:bop_factorial}, applying CAA to the fine-tuned
DINOv2 features results in only a minor decrease in average AR
(54.7$\rightarrow$54.5 for single-hypothesis and
57.9$\rightarrow$57.8 for five-hypothesis). This small drop is consistent
with the design goal of CAA, which reduces computation through adaptive
early exiting and therefore introduces a slight accuracy--efficiency trade-off.

\begin{table}[htbp]
\centering
\footnotesize
\caption{Controlled ablation of feature representation and aggregation strategy
under the BOP setting.}
\resizebox{0.48\textwidth}{!}{%
\begin{tabular}{ll|cc}
\hline
Feature & Aggregation & Single-hyp. AR & Five-hyp. AR \\
\hline
DINOv2    & Standard & 54.7 & 57.9 \\
DINOv2    & CAA      & 54.5 & 57.8 \\
Diffusion & Standard & {50.2} & {53.1} \\
Diffusion & CAA      & 55.5 & 58.3 \\
\hline
\end{tabular}%
}

\label{tab:bop_factorial}
\end{table}

In contrast, directly using the frozen diffusion features with standard
aggregation gives substantially lower performance (50.2 and 53.1 AR).
When combined with CAA, the performance increases to 55.5 and 58.3,
respectively. This indicates that the benefit of our method does not come
from simply replacing DINOv2 with frozen diffusion features; rather,
effective adaptive aggregation is important for exploiting the multi-scale
representations provided by the frozen diffusion backbone.

\subsection{Qualitative analysis}\label{sub:quali_any}
\subsubsection{Feature representation}\label{subsec:feat_rep}

\begin{figure}[htb]
\centering
\footnotesize
\subfloat{
\begin{minipage}[t]{0.05\linewidth} 
    \centering
    \vspace{1.2cm} 
    \rotatebox{90}{Unseen objects \quad \quad \quad \quad \quad \quad \quad  Seen objects } 
\end{minipage}%
\begin{minipage}[t]{0.470\linewidth}
    \centering
    \centerline{Template-pose}
    \includegraphics[width=1.55in]{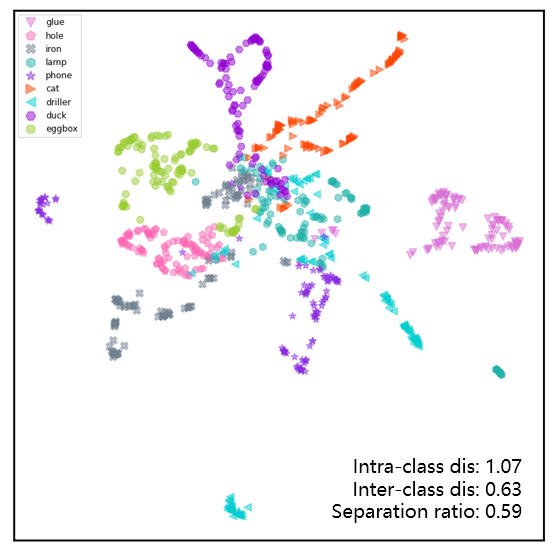}\\
    \includegraphics[width=1.55in]{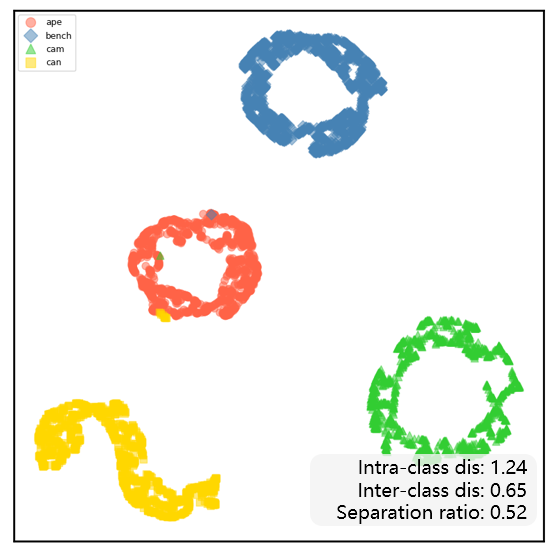}\\
\end{minipage}%
\begin{minipage}[t]{0.470\linewidth}
    \centering
    \centerline{Our method}
    \vspace{0.06cm} 
    \includegraphics[width=1.55in]{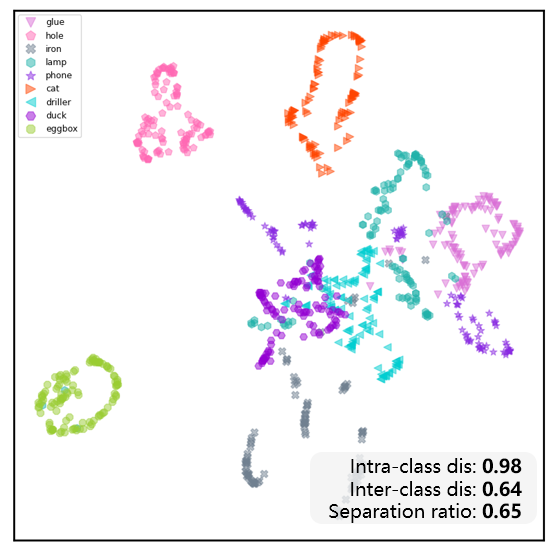}\\
    \includegraphics[width=1.55in]{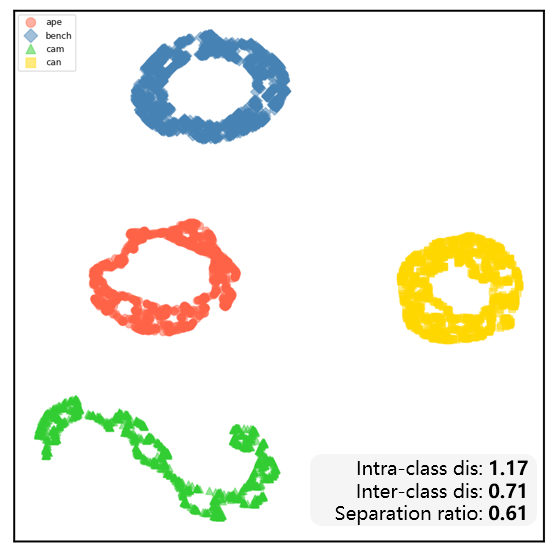}\\
\end{minipage}%
}
\centering
\caption{UMAP visualizations of feature representations after removing backgrounds using ground-truth object masks.}
\label{fig:feat_rep}
\end{figure}

In Fig. \ref{fig:feat_rep}, we present a comparison of the effectiveness of our method and the template-pose method in feature representation, utilizing UMAP~\citep{mcinnes2018umap} visualizations on the LM split \#1 dataset. The figure shows feature distributions for both seen and unseen objects. To reduce background interference, we use ground-truth object masks to provide a clearer visualization of the feature distributions for each object category.
\todooo{For quantitative comparison, we compute Euclidean feature distances in the original normalized feature space. Intra-class distance measures within-object pairwise distances, while inter-class distance measures distances between object centroids; their ratio defines the separation score.}

\todooo{For seen objects, our method achieves a tighter intra-class distribution,
with an average intra-class distance of {0.98}, compared to
{1.07} for the template-pose method. Additionally, it achieves an
inter-class distance of {0.64}, which is slightly higher than the
{0.63} achieved by the template-pose method. Consequently, our method
yields a higher separation ratio ({0.65 vs. 0.59}), indicating better
distinguishability between categories.}

\todooo{For unseen objects, our method also demonstrates {improved feature
separability, with a lower intra-class distance of 1.17 compared to 1.24 for
the template-pose method, and a higher inter-class distance of 0.71 compared
to 0.65}. Consequently, the separation ratio increases from
{0.52 to 0.61}. These results highlight that our method not only
improves feature representation for seen objects but also generalizes better
to unseen categories.}



\subsubsection{Pose discrimination}\label{subsec:pose_dis}
\begin{figure*}[ht]
\centering
\footnotesize
\rotatebox{90}{ \quad \quad \quad \; Our method \quad \quad \quad \quad \quad \quad Template-pose} 
\includegraphics[width=0.95\textwidth]{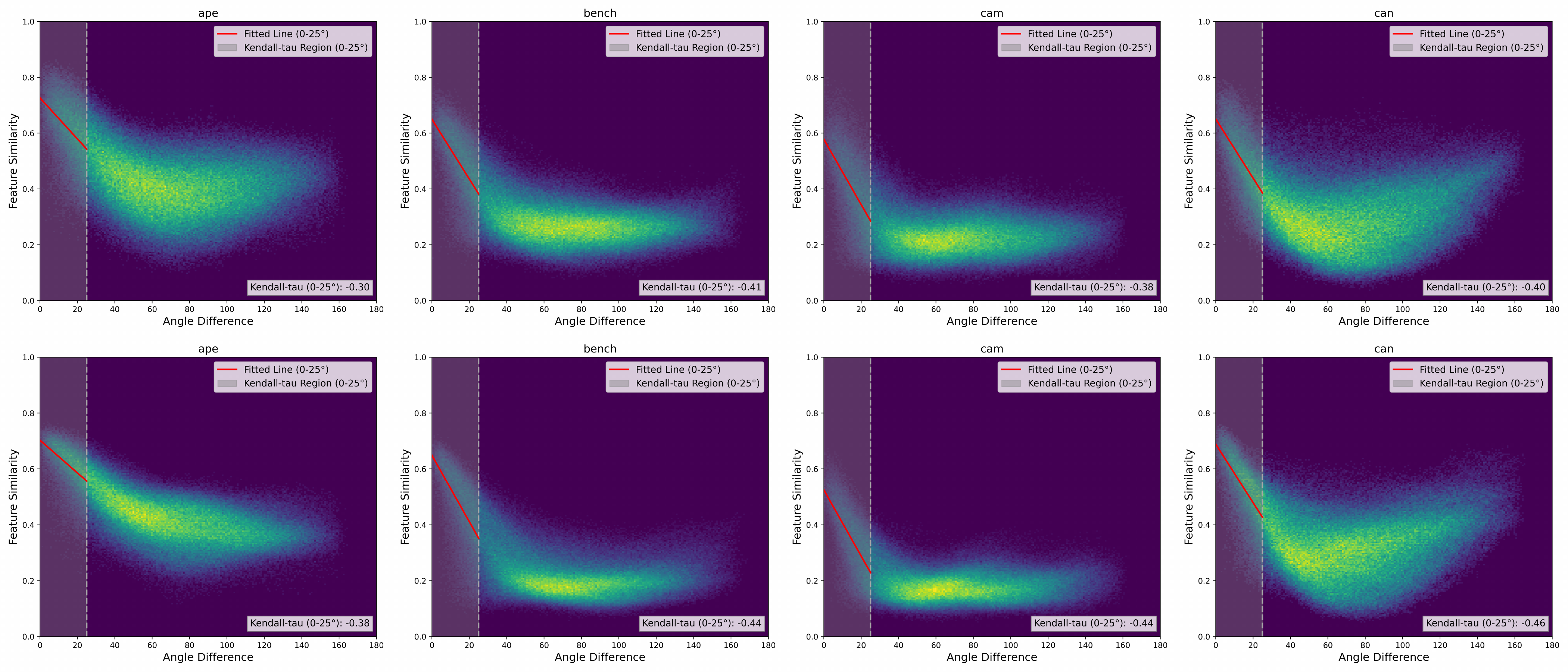}
\caption{Correlation between angle difference and feature similarity for template-pose (top) and our method (bottom) on the \emph{unseen} objects of LM split \#1 (ape, bench, can, cam). The shaded region (0-25°) highlights small pose differences, with the Kendall-tau coefficient used to assess pose discriminability.} 
\label{fig:pose_dis}
\end{figure*}

To further illustrate the superior discriminative power of our features in pose estimation tasks compared to existing methods, we analyze the correlation between angle difference (pose distance) and feature similarity. Figure \ref{fig:pose_dis} shows the results for our method and template-pose~\citep{nguyen2022templates} on the \emph{unseen} objects of LM split \#1.
The plots in Fig. \ref{fig:pose_dis} are generated by considering all possible pairs of real images and synthetic images for a given object. Ideally, these plots exhibit a negative correlation pattern, forming a diagonal trend close to -1, where feature similarity progressively decreases as angle difference increases. This ideal pattern demonstrates that the features effectively distinguish objects across various poses.

A key region in the figure highlights the behavior of feature similarity in small poses (0-25 °).
In an optimal scenario, feature similarity should concentrate within this region, indicating that the method achieves high stability and consistency under minimal pose variation. To further quantify this relationship, the Kendall-tau coefficient within the 0-25° range is computed to measure the correlation between feature similarity and pose difference. 
The Kendall-tau coefficient, which ranges from -1 to 1, reflects the strength of correlation. A value of -1 indicates perfect negative correlation (feature similarity decreases as pose difference increases). A value of 1 indicates perfect positive correlation, and 0 indicates no correlation.

Figure \ref{fig:pose_dis} illustrates that our method \todooo{consistently exhibits a more negative} Kendall-tau coefficient than the baseline within the 0-25° range, highlighting a stronger negative correlation between feature similarity and pose difference at small angles. This result demonstrates that our features are more stable and consistent under small pose variations, effectively capturing subtle feature differences with greater precision. 
Furthermore, in the highlighted region, our method exhibits a dense clustering of high feature similarity values at low pose differences. In contrast, template-pose~\citep{nguyen2022templates} shows a more dispersed distribution, indicating less effective differentiation across poses.
These findings emphasize that our method offers superior discriminative power and stability in pose estimation tasks compared to template-pose.

\subsubsection{Qualitative comparison of pose estimation}

\begin{figure*}[t]
\centering
\footnotesize
\rotatebox{90}{\quad\quad \normalsize\textbf{O-LM}}
\hspace{0.05in}
\subfloat{
    \begin{minipage}[t]{0.103\linewidth}
        \centering
        \includegraphics[width=0.7in]{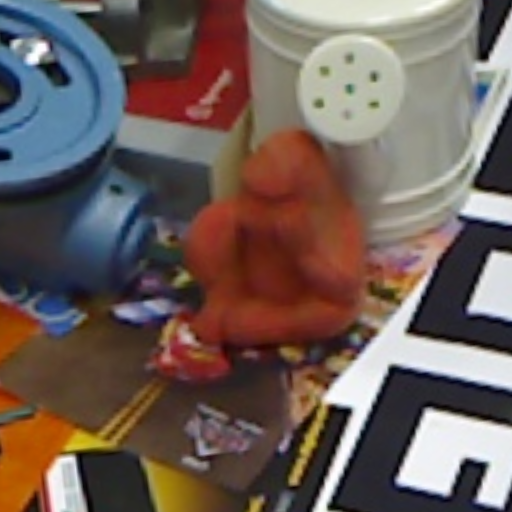}\\
        \vspace{0.005cm}
        \includegraphics[width=0.7in]{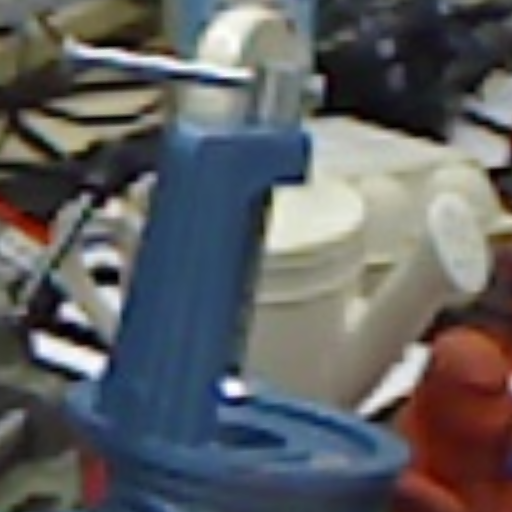}\\
        \vspace{0.005cm}
        \includegraphics[width=0.7in]{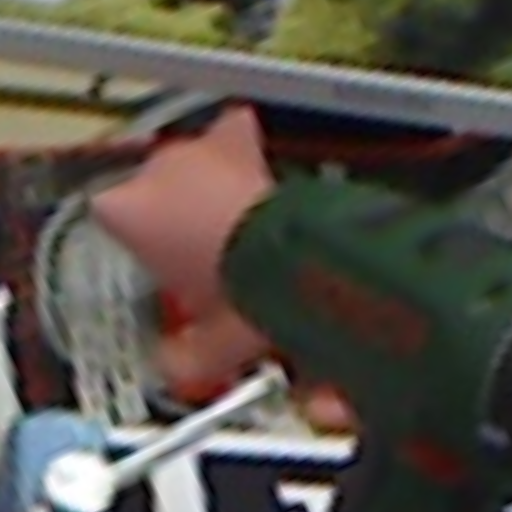}\\
        \vspace{0.005cm}
        \includegraphics[width=0.7in]{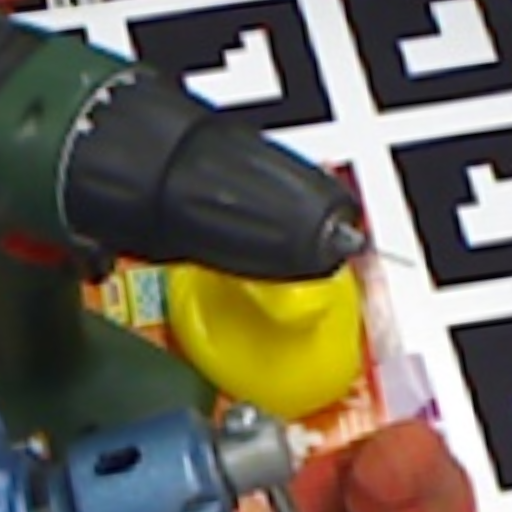}\\
        \centerline{Query}
    \end{minipage}%
        \begin{minipage}[t]{0.103\linewidth}
        \centering
        \includegraphics[width=0.7in]{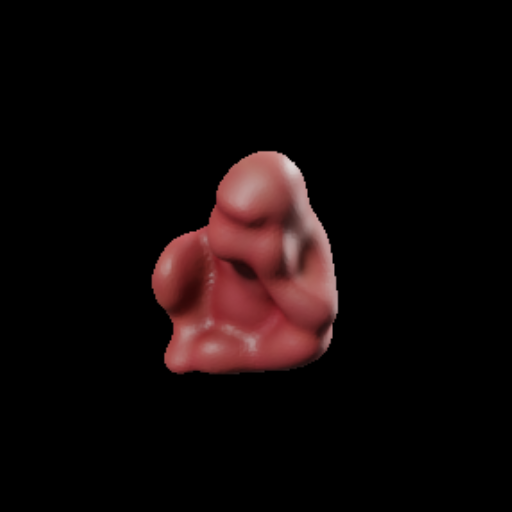}\\
        \vspace{0.005cm}
        \vspace{0.005cm}
        \includegraphics[width=0.7in]{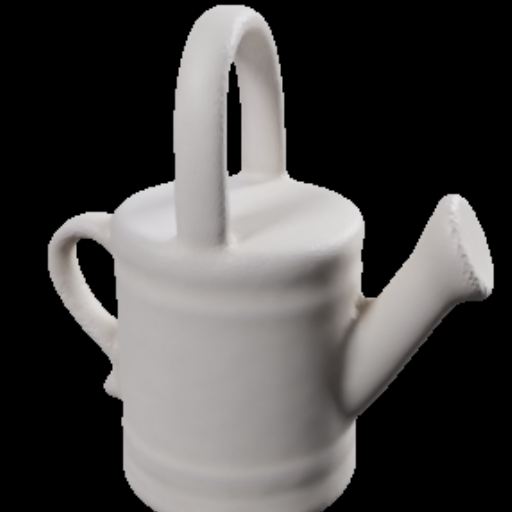}\\
        \vspace{0.005cm}
        \includegraphics[width=0.7in]{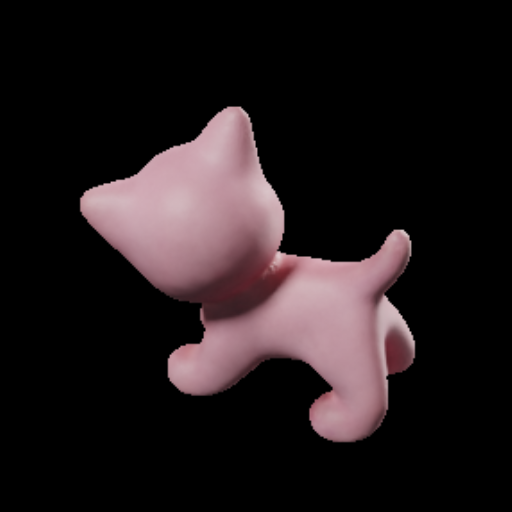}\\
        \vspace{0.005cm}
        \includegraphics[width=0.7in]{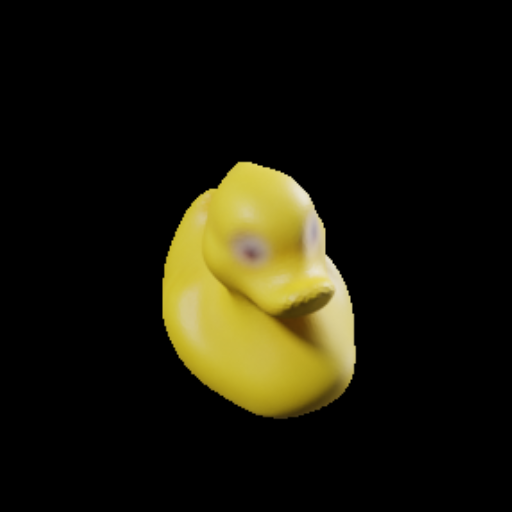}\\
        \centerline{GT}
    \end{minipage}%
    \begin{minipage}[t]{0.103\linewidth}
        \centering
        \includegraphics[width=0.7in]{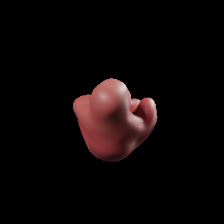}\\
        \vspace{0.005cm}
        \includegraphics[width=0.7in]{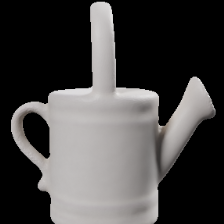}\\
        \vspace{0.005cm}
        \includegraphics[width=0.7in]{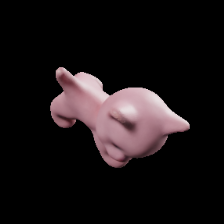}\\
        \vspace{0.005cm}
        \includegraphics[width=0.7in]{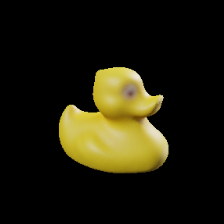}\\
        \centerline{Template-pose}
    \end{minipage}%
    \begin{minipage}[t]{0.103\linewidth}
        \centering
        \includegraphics[width=0.7in]{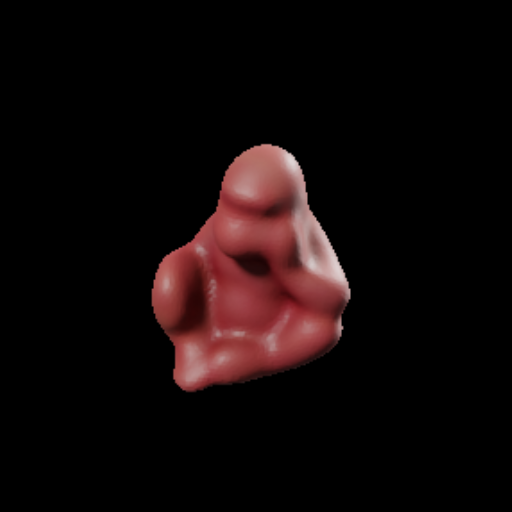}\\
        \vspace{0.005cm}
        \includegraphics[width=0.7in]{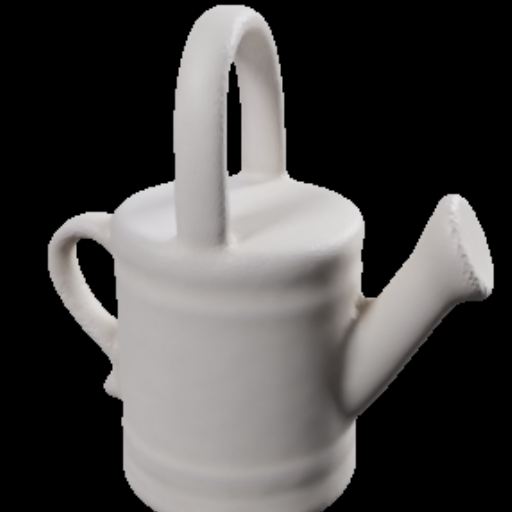}\\
        \vspace{0.005cm}
        \includegraphics[width=0.7in]{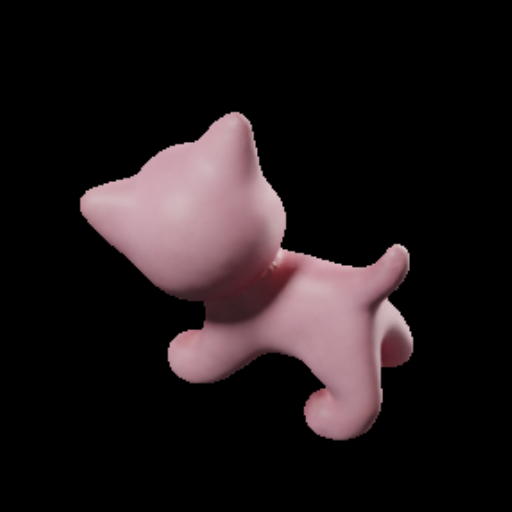}\\
        \vspace{0.005cm}
        \includegraphics[width=0.7in]{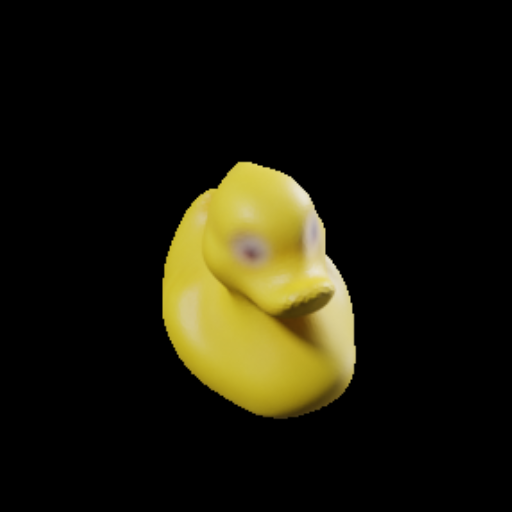}\\
        \centerline{Ours}
    \end{minipage}%
}
\hspace{0.25in}
\rotatebox{90}{\quad \ \normalsize\textbf{T-LESS}}
\hspace{0.05in}
\subfloat{
    \begin{minipage}[t]{0.103\linewidth}
        \centering
        \includegraphics[width=0.7in]{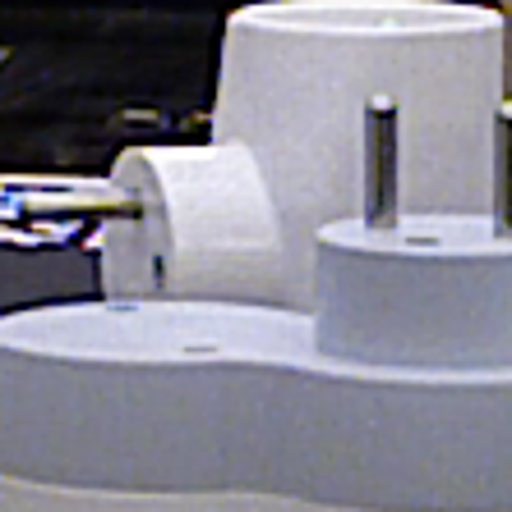}\\
        \vspace{0.005cm}
        \includegraphics[width=0.7in]{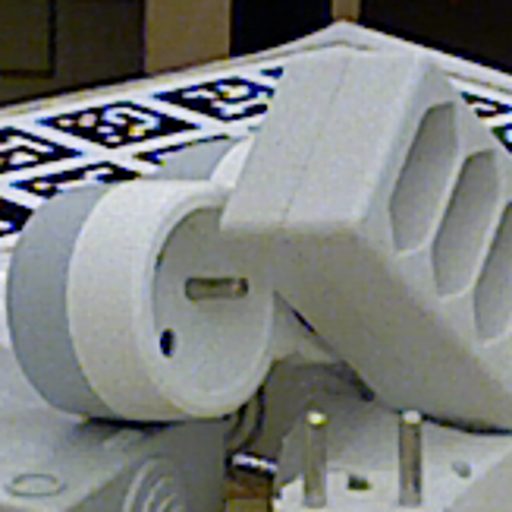}\\
        \vspace{0.005cm}
        \includegraphics[width=0.7in]{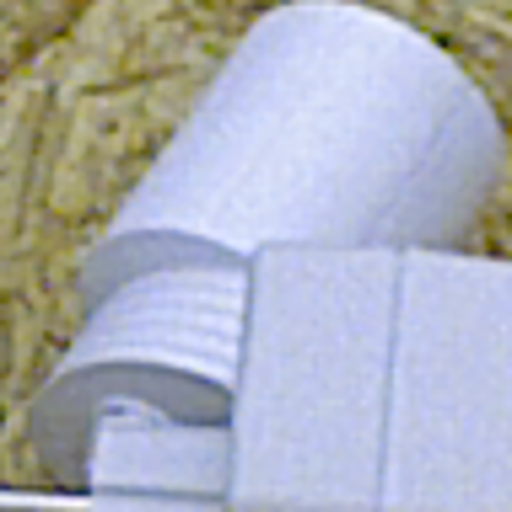}\\
        \vspace{0.005cm}
        \includegraphics[width=0.7in]{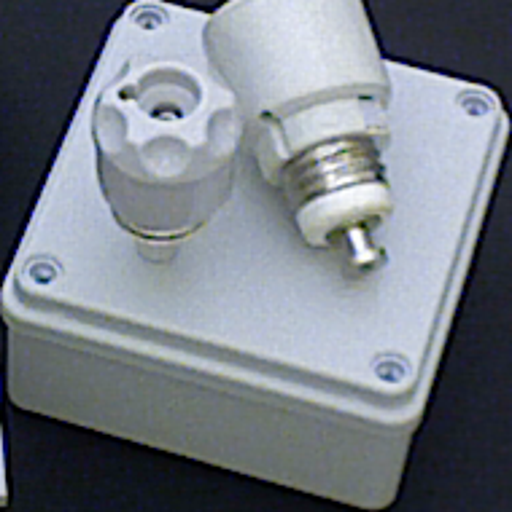}\\
        \centerline{Query}
    \end{minipage}%
    \begin{minipage}[t]{0.103\linewidth}
        \centering
        \includegraphics[width=0.7in]{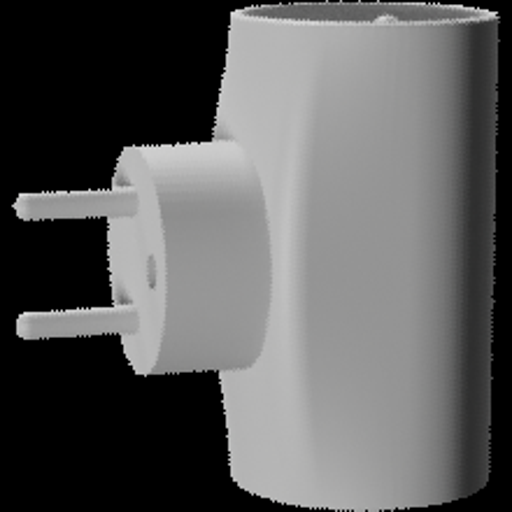}\\
        \vspace{0.005cm}
        \includegraphics[width=0.7in]{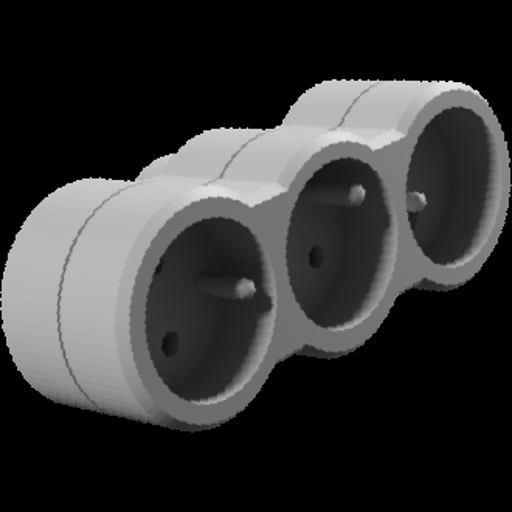}\\
        \vspace{0.005cm}
        \includegraphics[width=0.7in]{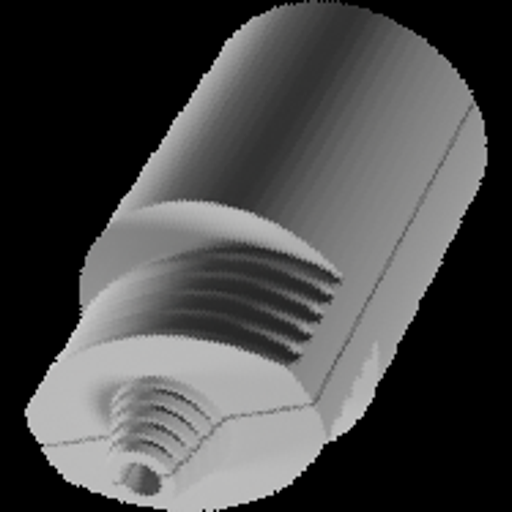}\\
        \vspace{0.005cm}
        \includegraphics[width=0.7in]{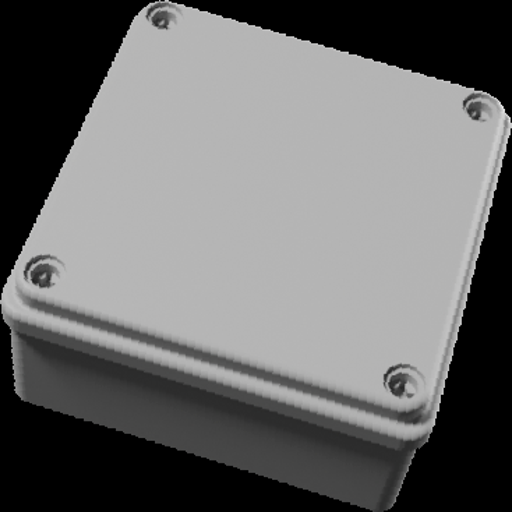}\\
        \centerline{GT}
    \end{minipage}%
    \begin{minipage}[t]{0.103\linewidth}
        \centering
        \includegraphics[width=0.7in]{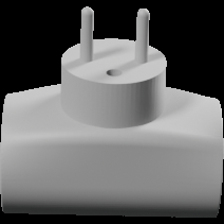}\\
        \vspace{0.005cm}
        \includegraphics[width=0.7in]{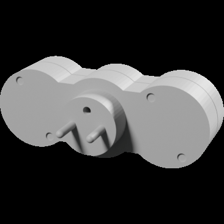}\\
        \vspace{0.005cm}
        \includegraphics[width=0.7in]{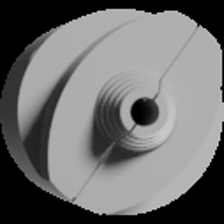}\\    
        \vspace{0.005cm}
        \includegraphics[width=0.7in]{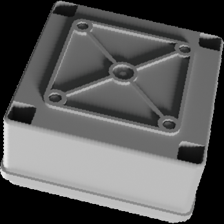}\\    
        \centerline{Template-pose}
    \end{minipage}%
    \begin{minipage}[t]{0.103\linewidth}
        \centering
        \includegraphics[width=0.7in]{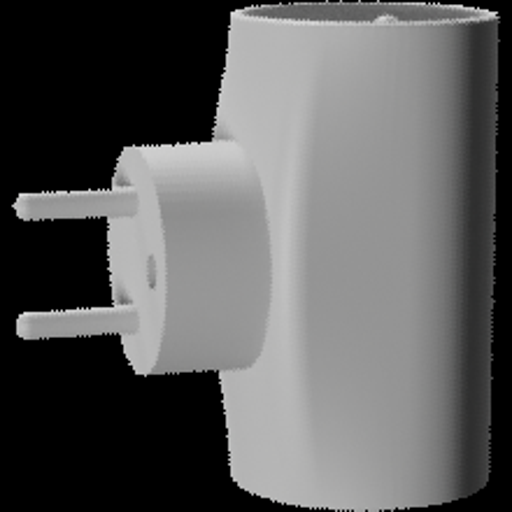}\\
        \vspace{0.005cm}
        \includegraphics[width=0.7in]{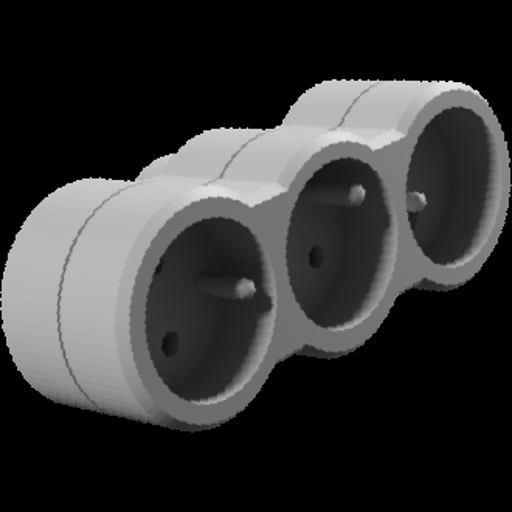}\\
        \vspace{0.005cm}
        \includegraphics[width=0.7in]{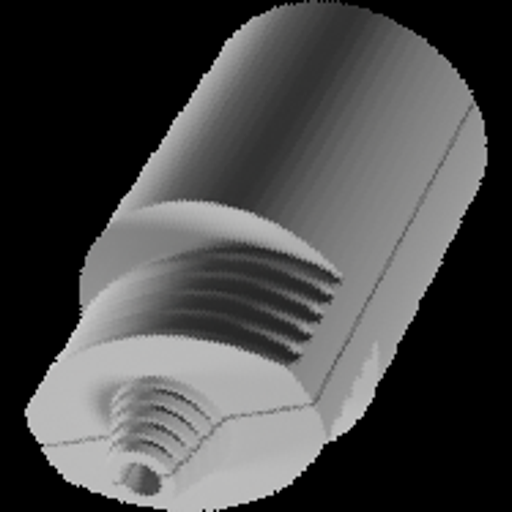}\\  
        \vspace{0.005cm}
        \includegraphics[width=0.7in]{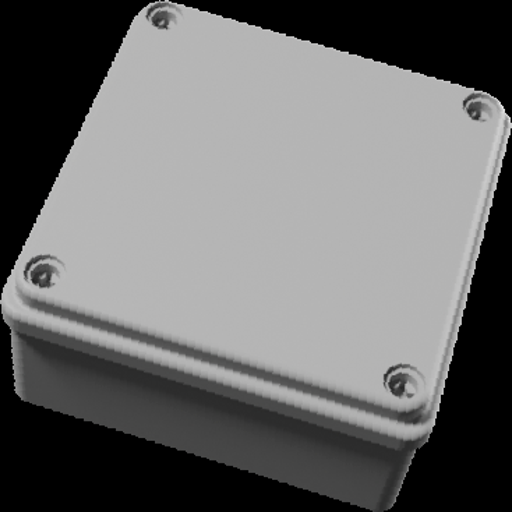}\\  
        \centerline{Ours}
    \end{minipage}%
}
\centering
\caption{{Qualitative results on unseen objects of O-LM (left) and T-LESS (right). }} 
\label{fig:vis}
\end{figure*}

To better demonstrate the effectiveness of our aggregation method, we provide several qualitative results in Fig. \ref{fig:vis}. We compare the estimated pose of our methods and the state-of-the-art method~\citep{nguyen2022templates}. We find that our method performs better in challenging scenarios, e.g. objects with occlusions in T-LESS, showing the stronger generalizability and discriminative capability of our method.

\subsection{\todoo{Parameter efficiency}} \label{sub:efficiency}
Table \ref{tab:param} displays the comparison of trainable parameter size  with template-pose~\citep{nguyen2022templates}. 
The proposed method outperforms template-pose with superior performance and approximately half the trainable parameters.

\begin{table}[htbp]
\centering
\caption{Comparison of trainable parameter sizes for different methods.}
\resizebox{0.48\textwidth}{!}{
\begin{tabular}{c|c|cccc}
\hline
 & Template-pose & VA & NA & CWA & CAA \\ \hline
Trainable Params (M) & 24.04 & 2.53 & 13.28 & 13.29 & 13.27\\ \hline
\end{tabular}
}

\label{tab:param}
\end{table}

\section{Conclusion}
\label{sec:conclusion}

In this study, we aim to achieve accurate and efficient pose estimation for objects. To tackle this challenge, we first analyze the issues leading to inaccurate pose estimation, particularly for unseen objects, and identify insufficient feature generalizability as the primary cause. To overcome this, we propose three novel aggregation networks specifically designed to effectively aggregate diffusion features, demonstrating superior generalizability for object pose estimation. Furthermore, to improve efficiency, we introduce a confidence-adaptive aggregation network that employs a dynamic feature selection mechanism, allowing our approach to achieve a better trade-off between computational cost and accuracy. We evaluate our method on various standard benchmark datasets, where it demonstrates superior performance and enhanced generalization to unseen objects compared to existing methods. We believe that our findings will contribute to further advancements in the field of object pose estimation.

In future work, we plan to focus on embedded deployment optimization to improve model efficiency on edge devices. Given the limited computational resources of these devices, we aim to leverage techniques such as model pruning and quantization to reduce model parameters and computational load, thereby significantly enhancing inference speed and energy efficiency while maintaining accuracy. Additionally, we will design a low-latency inference architecture to ensure that the model meets real-time application requirements in resource-constrained environments. These optimizations will help build a lightweight and efficient pose estimation model, suitable for deployment on smart devices in practical scenarios.


\noindent{\textbf{Code availability.} } All trained models and codes are publicly available on GitHub: \url{https://github.com/Tianfu18/diff-feats-pose}.

\noindent{\textbf{Data availability.} } The datasets used for evaluation during the current study are all publicly available. LINEMOD and Occlusion-LINEMOD are available at \url{https://drive.google.com/file/d/1XkQBt01nlfCbFuBsPMfSHlcNIzShn7e7/view}. TLESS is available at \url{https://drive.google.com/file/d/1d2GoswrnvcTlwFi_LWoCiy1uS5OkCiF1/view}. \todoo{The BOP Challenge datasets are available through the official BOP repository at \url{https://bop.felk.cvut.cz/datasets/}}.

{\footnotesize
	\bibliographystyle{plainnat}
	\bibliography{egbib}
}
\balance
\end{document}